%% file: main.tex
\documentclass[a4paper,11pt]{article}

\usepackage{jinstpub} 
\usepackage{multirow}
\usepackage{import}
\usepackage{subcaption}
\usepackage{xcolor}
\usepackage{siunitx}

\DeclareSIUnit{\sample}{S}
\DeclareSIUnit{\bit}{b}
\DeclareSIUnit{\equivalent}{ee}

\title{\boldmath
Investigating Resource-efficient Neutron/Gamma Classification ML Models Targeting eFPGAs}

\author[a,1]{Jyothisraj Johnson \note{Corresponding author.}}
\author[b]{Billy Boxer}
\author[a]{Tarun Prakash}
\author[a]{Carl Grace}
\author[a]{Peter Sorensen}
\author[b]{Mani Tripathi}

\affiliation[a]{Lawrence Berkeley National Laboratory (LBNL),\\ Berkeley, CA 94720-8099, USA}
\affiliation[b]{University of California, Davis, Department of Physics and Astronomy,\\ Davis, CA 95616-5270, USA}

\emailAdd{jyothisrajjohnson@lbl.gov}

\abstract{There has been considerable interest and resulting progress in implementing machine learning (ML) models in hardware over the last several years from the particle and nuclear physics communities. A big driver has been the release of the Python package, hls4ml, which has enabled porting models specified and trained using Python ML libraries to register transfer level (RTL) code. So far, the primary end targets have been commercial field-programmable gate arrays (FPGAs) or synthesized custom blocks on application specific integrated circuits (ASICs). However, recent developments in open-source embedded FPGA (eFPGA) frameworks now provide an alternate, more flexible pathway for implementing ML models in hardware. These customized eFPGA fabrics can be integrated as part of an overall chip design. In general, the decision between a fully custom, eFPGA, or commercial FPGA ML implementation will depend on the details of the end-use application. In this work, we explored the parameter space for eFPGA implementations of fully-connected neural network (fcNN) and boosted decision tree (BDT) models using the task of neutron/gamma classification with a specific focus on resource efficiency. We used data collected using an AmBe sealed source incident on Stilbene, which was optically coupled to an OnSemi J-series silicon photomultiplier (SiPM) to generate training and test data for this study. We investigated relevant input features and the effects of bit-resolution and sampling rate as well as trade-offs in hyperparameters for both ML architectures while tracking total resource usage. The performance metric used to track model performance was the calculated neutron efficiency at a gamma leakage of 10$^{-3}$. The results of the study will be used to aid the specification of an eFPGA fabric, which will be integrated as part of a test chip.}


\begin{document}
\maketitle
\flushbottom

\section{Introduction}
\label{sec:intro} 
\import{Text/}{intro.tex}

\section{The Python to RTL Workflow}
\label{sec:hls4ml} 
\import{Text/}{hls4ml.tex}

\section{Method}
\label{sec:method} 
\import{Text/}{method.tex}

\subsection{Resource Efficient BDT Models}
\label{sec:BDT}  
\import{Text/}{BDT.tex}

\subsection{Resource Efficient fcNN Models}
\label{sec:fcNN}  
\import{Text/}{fcNN.tex}

\section{Results}
\label{sec:results}  
\import{Text/}{Results.tex}

\subsection{BDT Results}
\label{sec:BDT_results}  
\import{Text/}{BDT_results.tex}

\subsection{fcNN Results}
\label{sec:fcNN_results}  
\import{Text/}{fcNN_results.tex}

\section{Conclusion}
\label{sec:conclusion}  
\import{Text/}{Conclusion.tex}

\acknowledgments {This work was supported in part by the U.S. Department of Energy under Contract No. DE-AC02-05CH11231. In addition, support was provided for collaborators at the University of California, Davis by award DE-NA0003996.}

\bibliography{main} 
\bibliographystyle{JHEP} 

\end{document}

%% file: Text/intro.tex

The initial release and continued development of the \emph{hls4ml} Python package over the past several years has enabled a complete Python to hardware pipeline for fast deployment of neural network (NN) based machine learning (ML) models. A detailed explanation of the \emph{hls4ml} workflow is given in \cite{fahim2021hls4ml} and \cite{duarte2018fast}. This package provides front-end converters to map models defined using standard Python ML libraries to an internal \emph{hls4ml} representation and then, after certain optimizations, directly outputs synthesizable C++ code that functionally describes the ML model. The package targets the input requirements for specific commercial high level synthesis (HLS) tools such as AMD/Xilinx Vivado HLS \cite{vivadohls} and Intel/Altera Quartus HLS \cite{intelhls} by using the appropriate C++ HLS libraries and pragmas/directives. Originally, support for boosted decision trees (BDTs) was integrated directly into \emph{hls4ml} along with the support for NN-based ML architectures \cite{Summers_2020}. Since then, the BDT workflow has been moved into \emph{conifer}, a complementary Python package released by the same collaboration. The original hardware target for these packages were commercial field-programmable gate array (FPGA) devices. However, over the years, the packages have been successfully used for fully custom ML model implementation on application specific integrated circuits (ASICs) \cite{di2021reconfigurable} and even more recently, embedded FPGA (eFPGA) fabrics designed using open-source frameworks \cite{gonski2024embedded}.

This latest hardware target was made possible by recent developments of open-source eFPGA frameworks, such as FABulous \cite{koch2021fabulous} and OpenFPGA \cite{tang2019openfpga}, which has removed the need to rely on expensive commercial vendors to integrate FPGA fabrics as part of an overall chip design. We intend to use FABulous, which allows for customization of the quantity and placement (location) of various primitive resources (also known as tiles) that will make up the fabric, chosen from those available in the framework. In addition, it provides the flexibility to implement custom tiles that can be integrated as part of an overall fabric. Overall, this technology allows for area-efficient solutions by specifying and sizing the eFPGA fabric for a given target application. However, there are several new considerations to using this technology. A detailed overview of the fabric generation workflow and these considerations are given in \cite{gonski2024embedded} and are summarized and expanded on here. Unlike a commercially available FPGA device, the availability of specific resource blocks, including memory resources, and their capabilities will depend on the framework and the technology node selected for physical implementation of the designed fabric. In addition, these resources must be explicitly included in the custom eFPGA fabric for them to be available. 

Achievable area-optimization for the eFPGA fabric will be limited by achievable routing densities and resource capabilities of the typically older process nodes used for the mixed-signal integrated chips targeting the applications under consideration \cite{chung2022shrink}. Explicitly, an eFPGA fabric implemented in \SI{28}{\nano\meter} versus \SI{180}{\nano\meter} will have very different capabilities for the same arrangement of primitive resource tiles defined using register transfer level (RTL) code. As a result, the exact latencies, resource usage, required maximum clock frequency, etc will be different for the same ML model depending on the process node selected for implementation. In general, the selected process node and the area taken up by the eFPGA fabric will directly affect the total cost associated with fabricating the final integrated chip (IC). Fabricating chips on newer process nodes will have higher costs per \unit{\milli\meter\squared}, but the overall area required to implement a specific eFPGA fabric will be lower. This is in contrast with cheaper per \unit{\milli\meter\squared} costs but much larger overall area required for the chip if using an older process node. Exact values and trade-offs will depend on specific foundries and their offerings, as well as the number (and type) of resources needed for the eFPGA fabric. Regardless, a much larger focus on optimizing resources to be included in the eFPGA fabric is required compared to commercial FPGAs. One example is provided in \cite{gonski2024embedded}, where limited resource availability in the final eFPGA fabric prevented higher performance BDT models from being deployed. 

A hardware-aware co-design methodology can aid in sparsifying ML models to target eFPGA deployments. For particle and nuclear physics applications, the eFPGA fabric is expected to be a part of an overall chip design, which will include dedicated front-end circuits to read out sensors, data converters, and digital back-ends to interface with data acquisition boards. By integrating data converter specifications into the ML model specification and training, we can trade-off the area/resource requirements of the eFPGA fabric with final converter specifications to deploy a specific ML model while maintaining the necessary level of performance. In addition, deployment of ML models on custom eFPGA fabrics means that we can also update the models as needed over the lifetime of the experiment to account for changes in detector conditions, different operations modes (i.e. calibration runs), etc on top of the benefit to ML-aided solutions to optimizing data to save for further offline analysis.

Because the number of resources (physical area) required to deploy ML models on eFPGAs will directly determine feasibility, this work explored the design parameter space (targeting eFPGA deployments) for two very basic ML algorithms: boosted decision tree (BDT) and fully connected neural network (fcNN) models. More specifically, we investigated feasibility for very sparse implementations of these basic ML models, which will translate to an area-efficient implementation for an appropriately sized custom eFPGA fabric to run these models on a future test chip. 

We used an older AMD/Xilinx Artix 7-series \cite{przybus2010xilinx} device (xc7a35ticsg324-1L) on an Arty development board as an initial stand-in for this exploration study. The exact resource usage numbers and other specifications achieved with this study using the stand-in commercial FPGA target will not directly translate to an open-source framework designed custom eFPGA fabric target. However, it is expected that the general trends for the models will hold and exact numbers will be off by some scaling factor. The commercial FPGA target has a total of 90 digital signal processing (DSP) blocks, 41.6k flip-flops (FFs), 20.8k lookup tables (LUTs) and the equivalent of 100 \SI{18}{\kilo\bit} random-access memory (RAM) blocks. Compared to the top of the line for the 7-series FPGAs, this is an approximately 20$\times$ reduction in resources available. Compared against the top of the line UltraScale+ family devices \cite{boppana2015ultrascale+}, this represents an approximately 60$\times$ reduction in available flip-flops and LUTs, 100$\times$ reduction for DSPs and 35$\times$ reduction for memory resources. The 7-series family devices are fabricated in a \SI{28}{\nano\meter} process node, representing a realistic upper bound comparison point on achievable performance and resource capabilities for the expected highest end node targeted for an actual custom eFPGA fabric. The amount of resources available on this commercial device is expected to be higher than for a reasonable custom FABulous eFPGA fabric, especially for the older process nodes. Thus, a main focus of this work was to target using only a small fraction of the overall resources available on this device when specifying a high performance ML model. We shall expand on the details in the next section.

To frame this feasibility study, the non-trivial task of neutron/gamma classification was used as an initial case study. We aim to use the results of this study to inform the specification of an actual eFPGA fabric to be implemented on the test chip, which will be able to deploy high performance ML models for this classification task. The task of neutron/gamma classification is either explicitly required or desired in a variety of radiation detection applications including neutron radiography \cite{hausladen2012fast} \cite{hausladen2013deployable}, associated particle imaging (API) \cite{heath2021development} and neutron scatter cameras \cite{braverman2018single} \cite{balajthy2023characterization} \cite{galindo2021design}. These applications typically happen in gamma dominated operating environments, necessitating very low false positive rates while still maintaining high neutron efficiency. The task is fundamentally multi-dimensional. Generally, specific plastic or organic crystalline scintillators with demonstrated inherent pulse shape discrimination (PSD) capability are used. This capability derives from the differences of a scintillator's singlet/triplet state activation with regards to the type of incident ionizing radiation \cite{Brooks_plastic}. However, other factors such as output scintillation wavelength and expected light yield also affect the overall discrimination power \cite{senoville2020neutron}. In addition, the time distribution of scintillation light produced is convolved with the silicon photomultiplier (SiPM) single photon response to produce an output electronic current pulse, introducing yet another dimension. The classification task itself is typically performed by directly or indirectly using differences in the waveform tail between neutron and gamma energy depositions. The standard discriminator used is a ratio between partial and total integration of the waveform.

The implementation of neutron/gamma classification ML models on commercial FPGAs is established. Previous examples of work done include but are not limited to \cite{fu2018artificial} \cite{michels2023real} \cite{astrain2021real} \cite{kaplan2019neutron} \cite{zhang2018real} \cite{yoon2022fast}. However, the combined advances in both open-source eFPGAs and translation of ML models to hardware now provide the basis for a viable pathway to a real-time, low-power and compact neutron/gamma classification method using ML models directly on a custom ASIC that integrates all required circuitry, including the eFPGA fabric. Thus, it is important to note that while this classification task is used as a case study for this work, a successful future eFPGA demonstration would open up new possibilities in implementing the required discrimination capability for the target applications. 

For this work, we use Stilbene \cite{SB_DS}, a widely utilized organic scintillator, and the OnSemi J-series 60035 SiPM \cite{onsemi:MICROJ−SERIES/D} to benchmark performance. While the SiPM single photon response is primarily set by details of the chosen SiPM, it is additionally shaped by any series resistance in the signal path and by the bandwidth of front-end electronics used \cite{acerbi2019understanding}. Although the details of specific front-end topologies are outside the scope of this work, we nevertheless indirectly explore the effects of front-end bandwidth in the study through data converter specifications.


Further details on using \emph{hls4ml} and \emph{conifer} to deploy ML models in hardware and the details of the workflow are given in Section \ref{sec:hls4ml}. In Section \ref{sec:method}, we provide a detailed overview of the study conducted in this work to explore the feasibility of a high performance eFPGA targeted ML model implementation, including setup and methodology. We introduce the test bed used for collecting data, provide an overview of the relevant hyperparameters for both types of ML architectures investigated in this study, and detail the metric used for evaluating model performance. In Section \ref{sec:results}, we look at the results of the study individually for both ML model architectures. Finally, in Section \ref{sec:conclusion}, we summarize these results and its significance for an eventual embedded field-programmable gate array (eFPGA) fabric implementation. 

%% file: Text/hls4ml.tex
The traditional FPGA implementation workflow uses RTL code, which specifies the intended behavioral functionality of a digital block, for design entry. From there, (vendor-specific) logic synthesis and place and route (PnR) software tools map the RTL code to resources such as LUTs, FFs, block RAM (BRAM), etc on a commercial FPGA target die. In the last two decades, commercial HLS software has matured and been adopted to allow for bypassing the requirement to write RTL code directly \cite{coussy2009introduction}. These tools use C/C++/SystemC/MATLAB code for design entry. The initial code provides an untimed, functional description of the intended digital block behavior and is then automatically mapped to a data flow graph (DFG), from which operations are scheduled based on capabilities of the target process node. The final output of the HLS software is (timed) RTL code that can be used in the standard FPGA workflow. 

As mentioned, the packages, \emph{hls4ml} and \emph{conifer} provide C++ input code for the HLS tools. In addition, they, respectively, allow specification of quantization levels for the biases/weights or thresholds of the fcNN/BDT model and provide support for pipelining operations. Because of this, the ML model can accept new inputs after waiting a set number of clock cycles (initiation interval), potentially before the final prediction for the previous inference is available (achievable latency). This capability is crucial for potentially replacing neutron/gamma classification circuitry of a large number of individual channels with a single pipelined ML model implemented on an eFPGA fabric. The \emph{hls4ml} package is flexible enough that it provides several user-configurable settings, which can be set as needed to explore trade-offs in latency, throughput, power and resource usage for the classification task at hand \cite{duarte2018fast}. In comparison, \emph{conifer} does not provide as extensive configurability, but the design parameter space for BDTs is also simpler \cite{Summers_2020}.

To explore the full parameter space, we had to pick specific Python libraries to specify and train the ML models. While both \emph{scikit-learn} \cite{scikit-learn}  and \emph{XGBoost} \cite{chen2016xgboost} have Python libraries allowing for the specification and training of BDTs, it is generally accepted that \emph{XGBoost's} version results in both better performance and faster training \cite{chen2016xgboost}. Furthermore, preliminary checks of resource usage for similarly set hyperparameter values between BDT models specified using the two libraries indicate very similar resource usage values. For these reasons, we used \emph{XGBoost's} BDT model for this work. For fcNNs, several Python libraries are also available. Currently, both \emph{Pytorch} \cite{ansel2024pytorch} and \emph{TensorFlow} \cite{tensorflow2015-whitepaper}, which has integrated \emph{Keras} \cite{chollet2015keras}, support performing quantization aware training (QAT). However, \emph{hls4ml} primarily supports QAT, at present, through the \emph{QKeras} front-end \cite{coelho2020ultra}. Taking into consideration the importance of this step in the overall aim of resource-efficient ML eFPGA implementations \cite{fahim2021hls4ml}, we have used \emph{QKeras} for specifying the fcNN models. 

\begin{figure*}[htb!]
    \centering
    \includegraphics[width=0.85\textwidth]{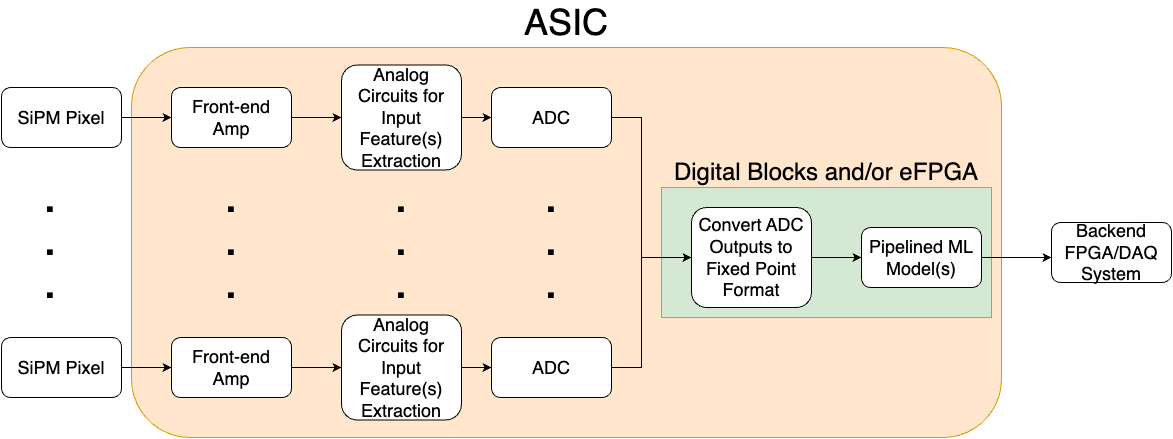}
    \includegraphics[width=0.85\textwidth]{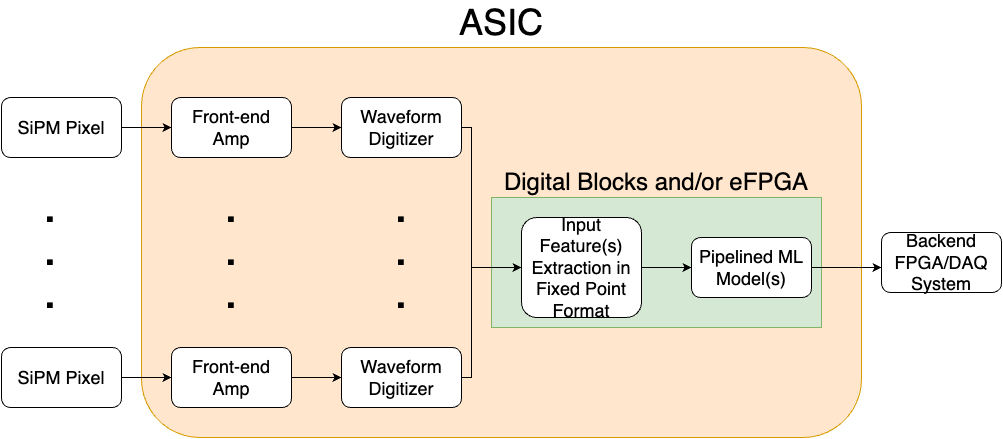}
\caption{A system diagram illustrating the general signal progression when analog circuits are used to calculate input feature values with ADCs digitizing those values on top and when a waveform digitizer and digital signal processing block are used on the bottom.}
    \label{fig:signal_chain_both}
\end{figure*}

Implementing ML models in hardware requires consideration of several new factors that are generally not necessary for a software implementation. The choice of input features to use in the ML model will depend on the ability to extract these features in real-time. Depending on the exact input features used, analog circuits or digital signal processing algorithms can be used to extract input feature values for a given event. The diagrams in Figure \ref{fig:signal_chain_both} present a generic system signal progression diagram comparing both scenarios. The exact input features required and a decision on if the ability to redefine the input features used is desired will guide such a decision. A re-programmable eFPGA fabric will need to be sufficiently specified to enable such an ability. In either case, the list of circuitry includes necessary front-end electronics for direct SiPM read out, analog-to-digital converters (ADCs) for either waveform or individual input feature digitization and time-to-digital converters (TDCs) for capturing timing information in addition to dedicated digital blocks and/or an appropriately specified eFPGA fabric. A more detailed look at differences between using a waveform digitizer or analog circuits is given in Section \ref{sec:method} using a comparative look at the performance of the standalone charge comparison pulse shape discrimination method. Regardless, it is clear that the specifications of the ADC in both scenarios will directly set the best possible input feature value quantization. In other words, the ADC specification can play a significant role on achievable model performance. To account for this, we chose to fold in the ADC bit-resolution and sampling rate as parameters in our analysis to understand possible trade-offs of ADC requirements for a future test chip (i.e. perform a hardware aware co-design of the fcNN/BDT models).

\begin{figure*}[htb!]
    \centering
    \includegraphics[width=0.6\textwidth]{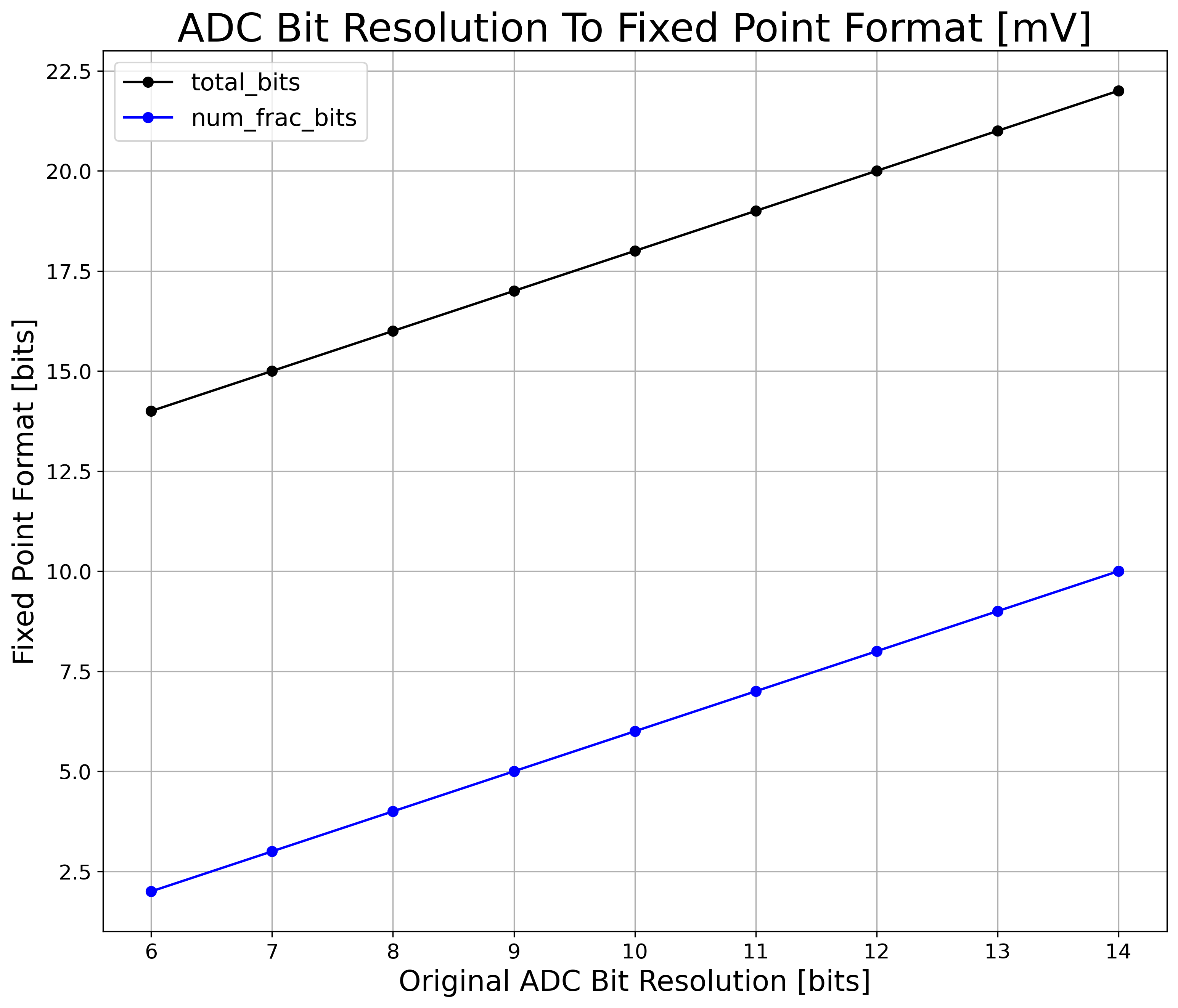}
    \includegraphics[width=0.6\textwidth]{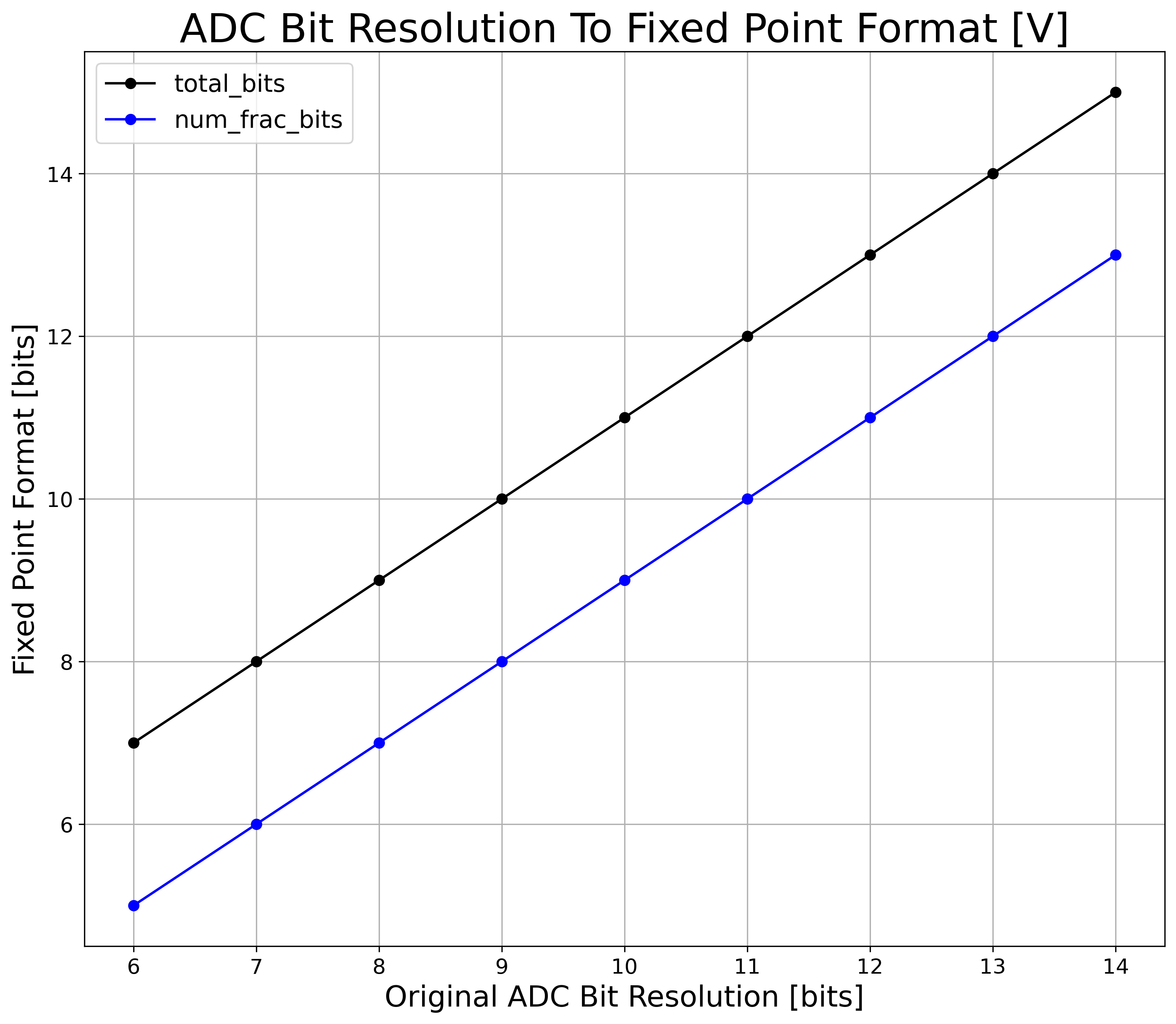}
\caption{The relationship between ADC bit resolution and the number of bits required in fixed point format to represent values in a base unit of \unit{\volt} or \unit{\mV} over a \SI{2}{\volt} full-scale range. The top figure shows the mapping for \unit{\mV} and the bottom, for \unit{\volt}.}
    \label{fig:adcbitres}
\end{figure*}

The quantization of the input features plays a significant role in lowering overall resource usage. In the extreme case, keeping native 32-bit resolution leads to larger resource usage for not much better performance \cite{duarte2018fast}. For our end-use target application(s), as mentioned, we anticipate having on-chip ADCs. The specifications of the ADC's bit resolution and full scale range will set the upper bound on the input feature quantization. The process of assigning ADC counts a physical value allows for selecting the base unit, which plays a non-trivial role in the required number of bits per input feature. For example, using millivolts (\unit{\mV}) or Volts (\unit{\volt}) for storing the value of a pulse's peak amplitude is straightforward and interchangeable in software. However, both \emph{hls4ml} and \emph{conifer} use fixed point format to represent values for input feature. Thus, the number of integer and fractional bits explicitly need to be set. An additional bit will be used for the sign. Figure \ref{fig:adcbitres} explicitly demonstrates the difference in input word lengths required to exactly represent a value over a \SI{2}{\volt} full-scale range depending on if \unit{\mV} or \unit{\volt} is the base unit. For this full-scale range, the main difference is due to an upfront cost in needing 11 integer bits to represent the maximum integer component when using \unit{\mV} compared to a single bit when using \unit{\volt}. The number of fractional bits scale with ADC bit-resolution. We expand on this point and the assumptions used in this study in Section \ref{sec:method}.

The overall requirements on neutron/gamma classification must also take into account the expected range of operating environments for the end-use applications. More specifically, the expected range of both gamma and neutron rates encountered will directly affect the absolute efficiency requirement on the neutron/gamma classification circuitry. We can tolerate larger gamma leakage values if the gamma rates will be less than or equal to neutron rates in the range of expected operating environments. However, in gamma dominated operating environments, much lower gamma leakage values are required. In this work, we assume operation in a gamma-dominated environment and define the performance metric used to assess performance of trained machine learning (ML) models as the neutron efficiency at a gamma leakage of 10$^{-3}$. This is not an absolute lower bound on expected gamma leakage requirements across all potential end-use applications. Rather, this is due to a statistical limitation from the size of the training data set used for the study as explained further in Section \ref{sec:method}.

%% file: Text/method.tex
\begin{figure}[tb]
    \centering
    \includegraphics[width=0.98\textwidth]{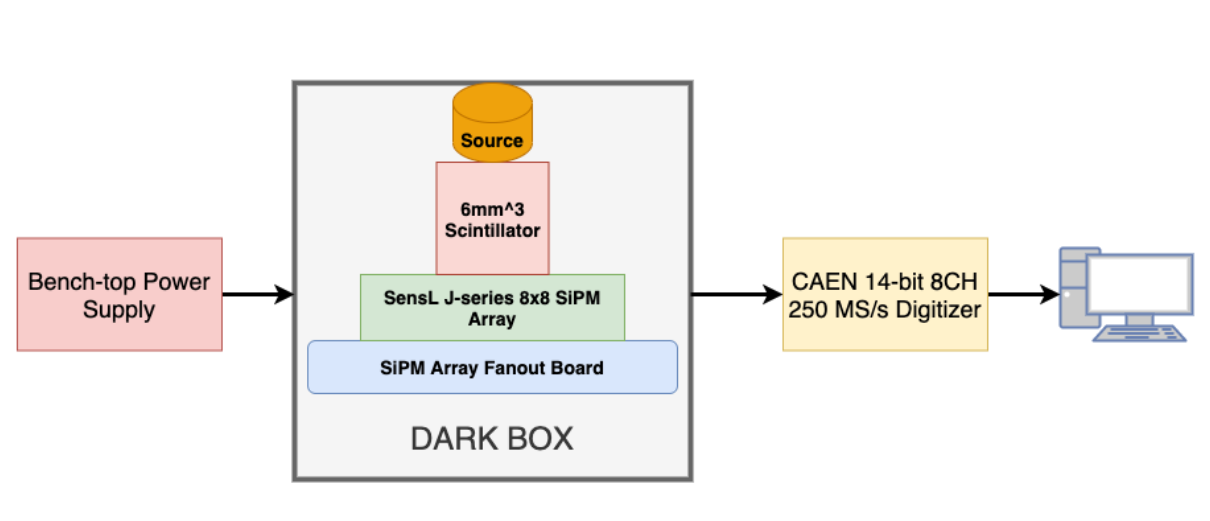}
    \caption{Diagram of the bench top test bed used to acquire the waveform dataset used in this work. Figure is replicated from \cite{Boxer_2023}.}
    \label{fig:benchtop_setup}
\end{figure}

The neutron and gamma waveform data set used in this work was acquired with the same benchtop test bed described in detail in \cite{Boxer_2023}. We briefly summarize the test bed again here. Figure \ref{fig:benchtop_setup}, replicated from \cite{Boxer_2023}, provides a system diagram of the setup. A custom SiPM array fanout board was used to read out selected channels of an 8$\times$8 array of OnSemi J-series 60035 SiPMs \cite{onsemi:MICROJ−SERIES/D}. Each standard, resistively coupled output (SOUT) signal is connected in series with a \SI{50}{\ohm} resistor, which provides current-to-voltage conversion and appropriate impedance matching with the characteristic impedance of the coaxial connectors that the signal is routed to. Importantly, it also dominates the single photon response. The additional series resistance increases the decay constant associated with the recharging of the SiPM's single photon avalanche diode (SPAD) microcells after receiving a hit. The waveform tails extend over a time period > \SI{1}{\micro\second} as a result. This fact should be kept in mind when interpreting and extrapolating any of the results of this study. Each fast, capacitively coupled output (FOUT) signal, already a voltage output, is connected to a balun transformer that provides the correct interface to the coaxial connector that brings the signal off the board. Signals are digitized using a commercial 14-bit, \SI{250}{\mega\sample\per\second} CAEN DT5725 digitizer with data collected using a computer interface. Specifically, the digitized waveforms were collected for a \SI{6}{\milli\meter} cube of Stilbene optically coupled to a single SiPM in the array with AmBe used as a mixed neutron/gamma emitter. 

\begin{figure}[tb]
    \centering
    \includegraphics[width=0.7\textwidth]{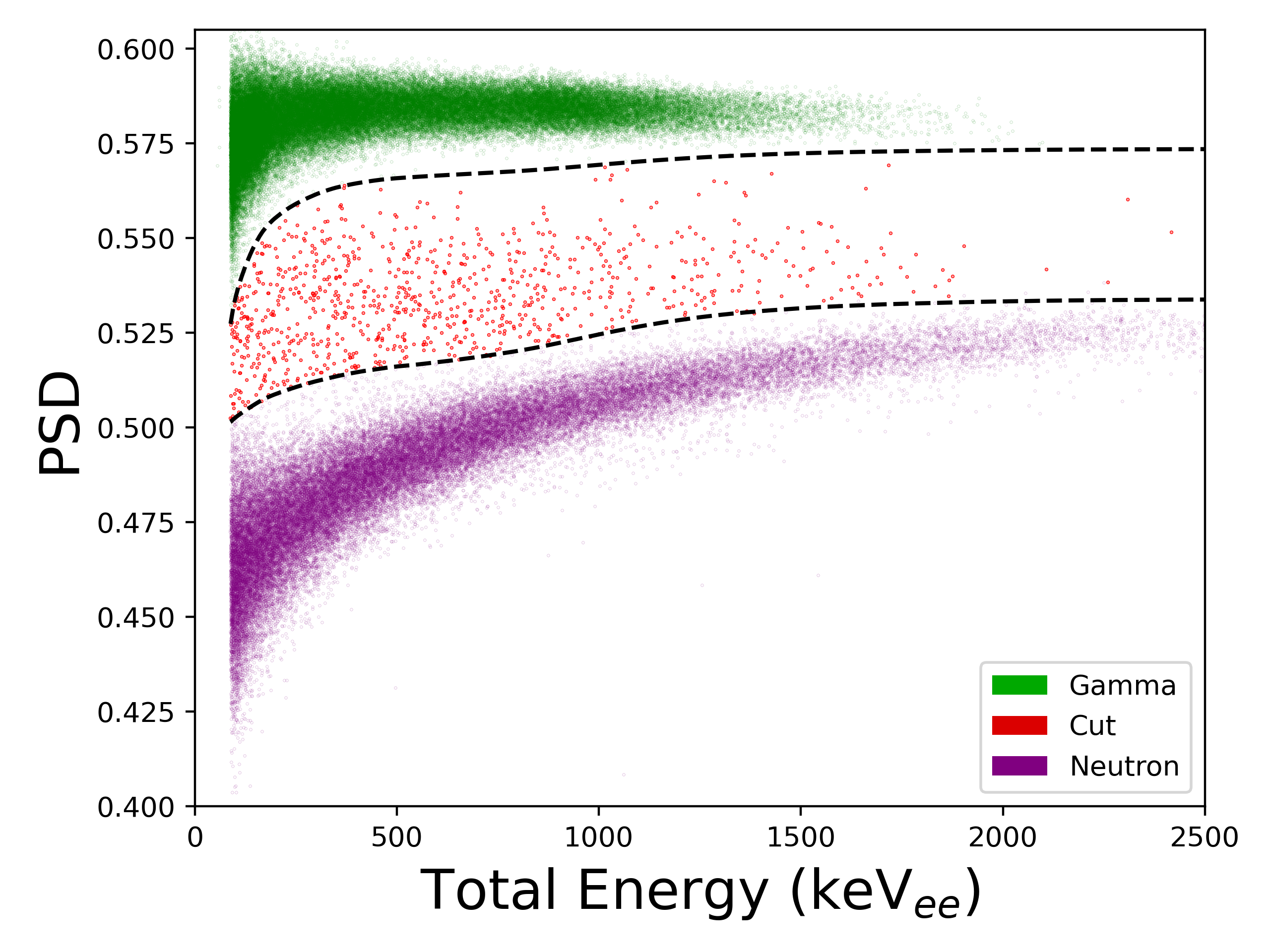}
    \caption{The neutron and gamma events used for this study are plotted as a function of PSD ratio vs. total deposition energy. A partial integration window of \SI{128}{\nano\second} and total integration window of \SI{1500}{\nano\second} is used. An energy cut at \SI{90}{\kilo\electronvolt}ee is included. Afterwards, upper and lower threshold cuts were defined to exclude events in between the neutron and gamma bands to maintain purity of the truth data set.}
    \label{fig:PSD_cut}
\end{figure}

In total, 80k gamma and 49k neutron waveforms were selected from a full 30 million event data set with an energy cut selecting only events with > \SI{90}{\kilo\electronvolt}ee energy depositions. As stated, the goal of this study was to demonstrate feasibility of high performance ML model deployments on an eFPGA fabric, not necessarily to improve classification efficiency at low energy depositions. The vast majority of events in the full data set correspond to the \SI{60}{\kilo\electronvolt} Am-241 X-ray emission. Truth labels were assigned by performing a PSD ratio cut, with total integration set to \SI{1500}{\nano\second} and partial integration set to \SI{128}{\nano\second}, in the standard PSD ratio vs energy parameter space. Figure \ref{fig:PSD_cut} shows these cuts and the resulting distribution of gamma and neutron events. The inversion of the neutron and gamma bands in the figure result from an inversion in the definition of PSD used. Instead of a tail-over-total, we performed a partial-over-total ratio as defined in Equation \eqref{eq:PSD} with the start of the integration defined as $t_0$, the partial integration ending at $t_1$ and the total at $t_2$:

\begin{equation} \label{eq:PSD}
\mathrm{PSD}=\frac{\int_{t_0}^{t_1} \mathrm{Waveform(}t\mathrm{)}\; \it{dt}}{\int_{t_0}^{t_2} \mathrm{Waveform(}t\mathrm{)}\;\textit{dt}}.
\end{equation} 

Further details for this reasoning are found in \cite{Boxer_2023}. The 130k total events in the dataset confines the  gamma leakage metric to 10$^{-3}$. Going a magnitude lower means that the obtained neutron efficiency values will depend on the ability to classify only a few events from the full dataset, once split into training and truth/validation subsets. This would mean an added level of statistical uncertainty, which does not map to a true physical limit. 

\begin{figure}[tb]
    \centering
    \includegraphics[width=0.6\textwidth]{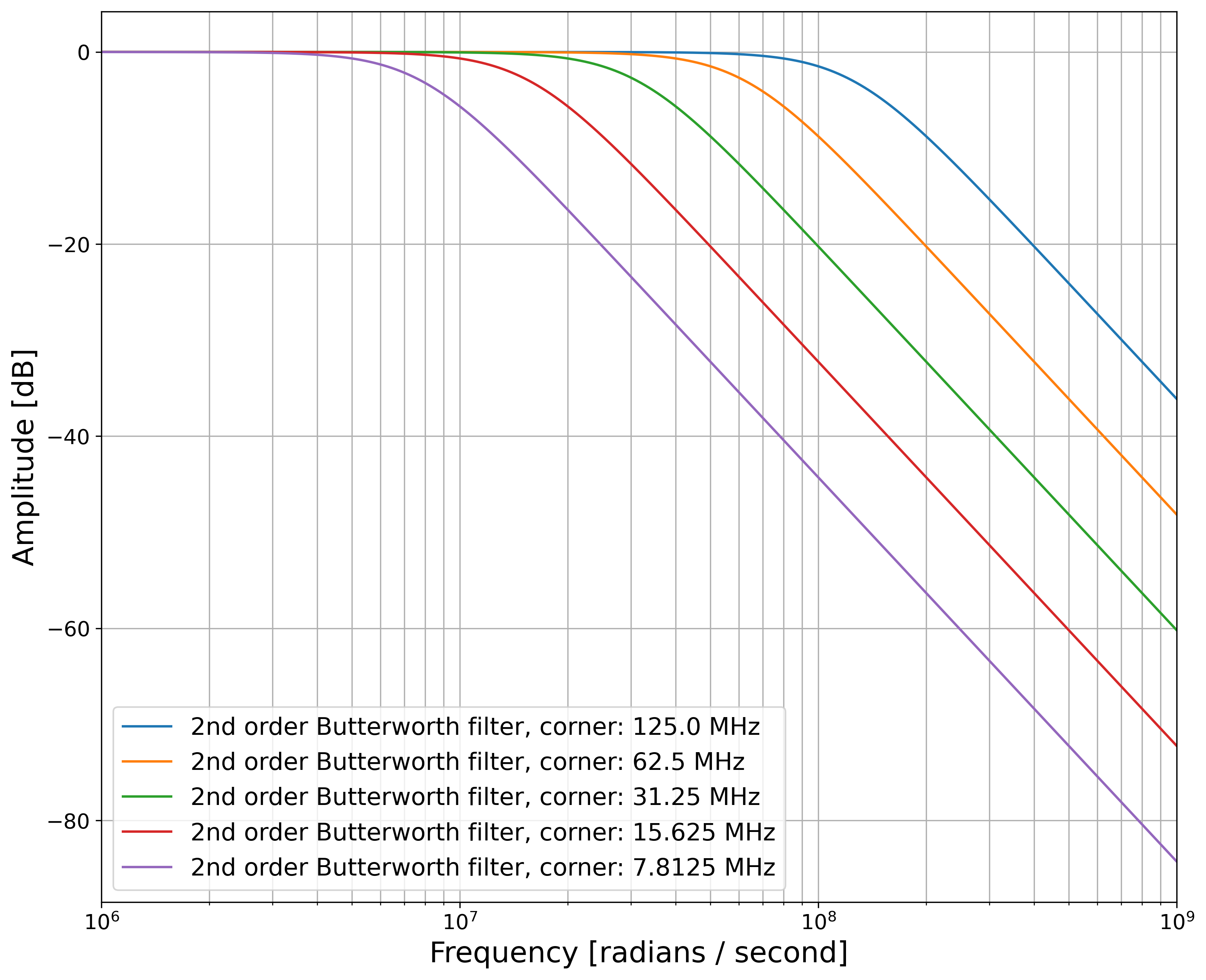}
    \caption{The frequency response of the 2nd-order Butterworth filters used to re-filter the waveforms in software. Corner frequencies were specified corresponding to a 1$\times$, 2$\times$, 4$\times$, 8$\times$, and 16$\times$ downscaling of the actual \SI{125}{\mega\hertz} anti-aliasing filter used by the CAEN digitizer.}
    \label{fig:bodeplot}
\end{figure}

\begin{figure}[tb]
    \centering
    \includegraphics[width=0.8\textwidth]{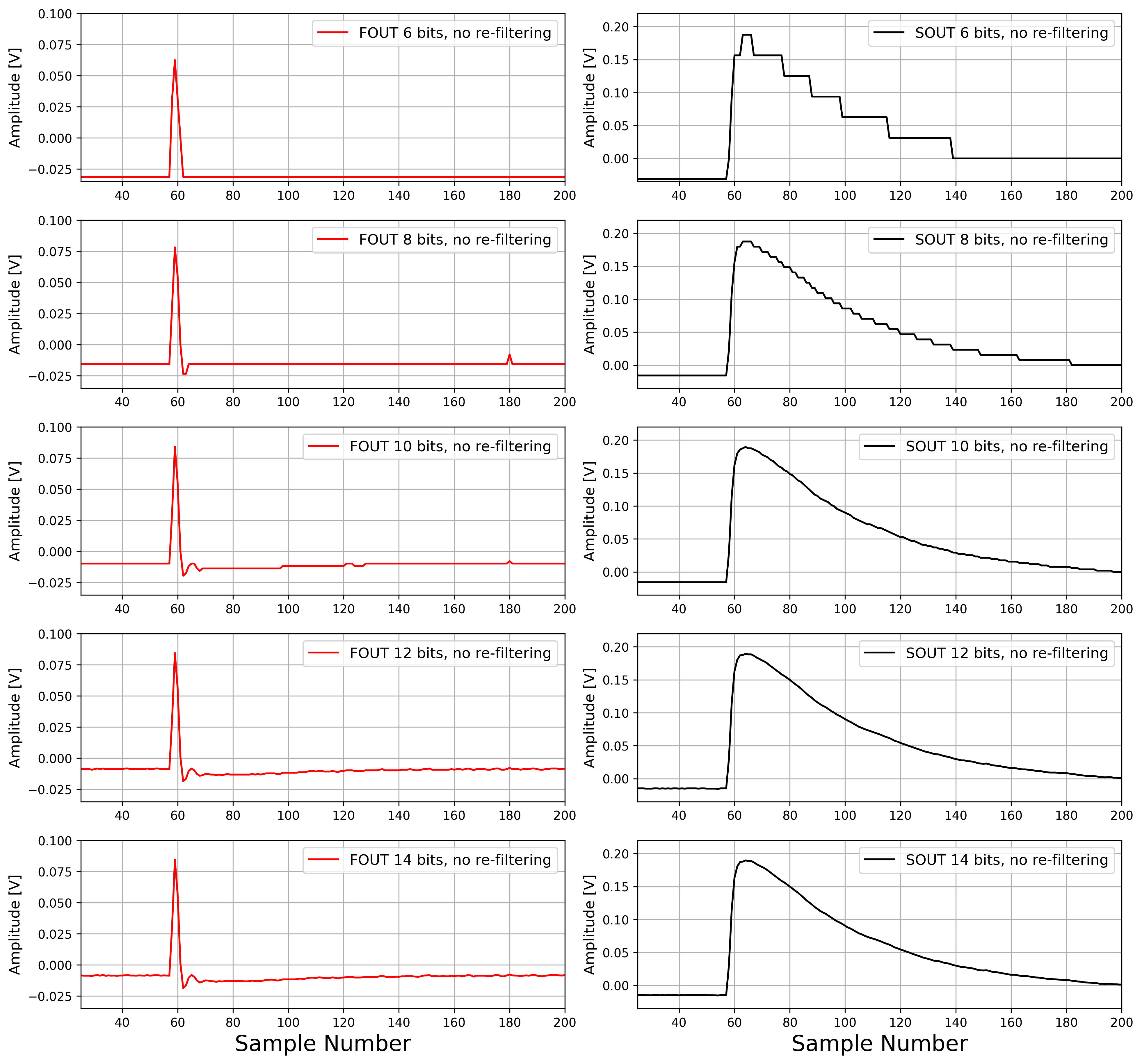}
    \caption{Example FOUT and SOUT waveforms re-digitized in software to various bit-resolutions without any applied software re-filtering. Each sample corresponds to the original \SI{250}{\mega\sample\per\second} rate.}
    \label{fig:redigitized_waveforms}
\end{figure}

Although the original waveforms were digitized with 14-bit resolution over \SI{2}{\volt} full-scale and centered at \SI{1}{\volt} at a sampling rate of \SI{250}{\mega\sample\per\second}, as mentioned, we wanted to understand the effect of ADC bit-resolution and sampling rate on the performance of trained BDT and fcNN models. To do so, we can perform re-digitization, digital filtering and decimation of the waveform samples in software in order to emulate a range of bit-resolutions, front-end bandwidths and effective sampling rates. In total, we considered bit-resolutions from eight to the original 14 bits and sampling rates between \SI{15.625}{\mega\sample\per\second} (decimation by a factor of 16) to the original \SI{250}{\mega\sample\per\second}. In addition, re-filtering of the waveforms without downsampling was also performed to take a closer look at the expected degradation in performance of the benchmark charge comparison PSD method. It is important to note that the measured thermal noise of the test bed was \SI{180}{\micro\volt}, which limited effective resolution to around 12 bits. 

The CAEN digitizer uses a second-order, \SI{125}{\mega\hertz} anti-aliasing filter. We emulated the filter in software using a second-order Butterworth filter implemented with the \emph{SciPy} \cite{SciPy} Python package. The critical frequency (corner frequency) of the filter was varied to be half the effective targeted sampling rate to match the physical digitizer's Nyquist configuration. The corner frequency, also known as the \SI{3}{\decibel} point, represents the value at which the filter attenuates the input voltage signal by a factor of \(\frac{1}{\sqrt{2}}\). Typically, the visual logarithmic representation of the frequency response is shown using a Bode plot \cite{haidekker2020linear}. Figure \ref{fig:bodeplot} shows this combined Bode plot of the various filter responses for the different targeted sampling rates. Five effective sampling rates were emulated by downsampling (every Nth sample kept) the original waveform samples by N, where N is the ratio of the original \SI{250}{\mega\sample\per\second} sampling rate divided by the targeted sampling rate. In total, the parameter space encompassed by these two variables includes 35 unique pairs of effective bit-resolution and sampling rate. Figure \ref{fig:redigitized_waveforms} shows a representative FOUT and SOUT neutron waveform at the original 14-bit resolution and at a selection of other bit-resolutions down to 6 bits. No software re-filtering of the waveforms was performed for this figure.

\begin{figure}[tb]
    \centering
    \includegraphics[width=0.8\textwidth]{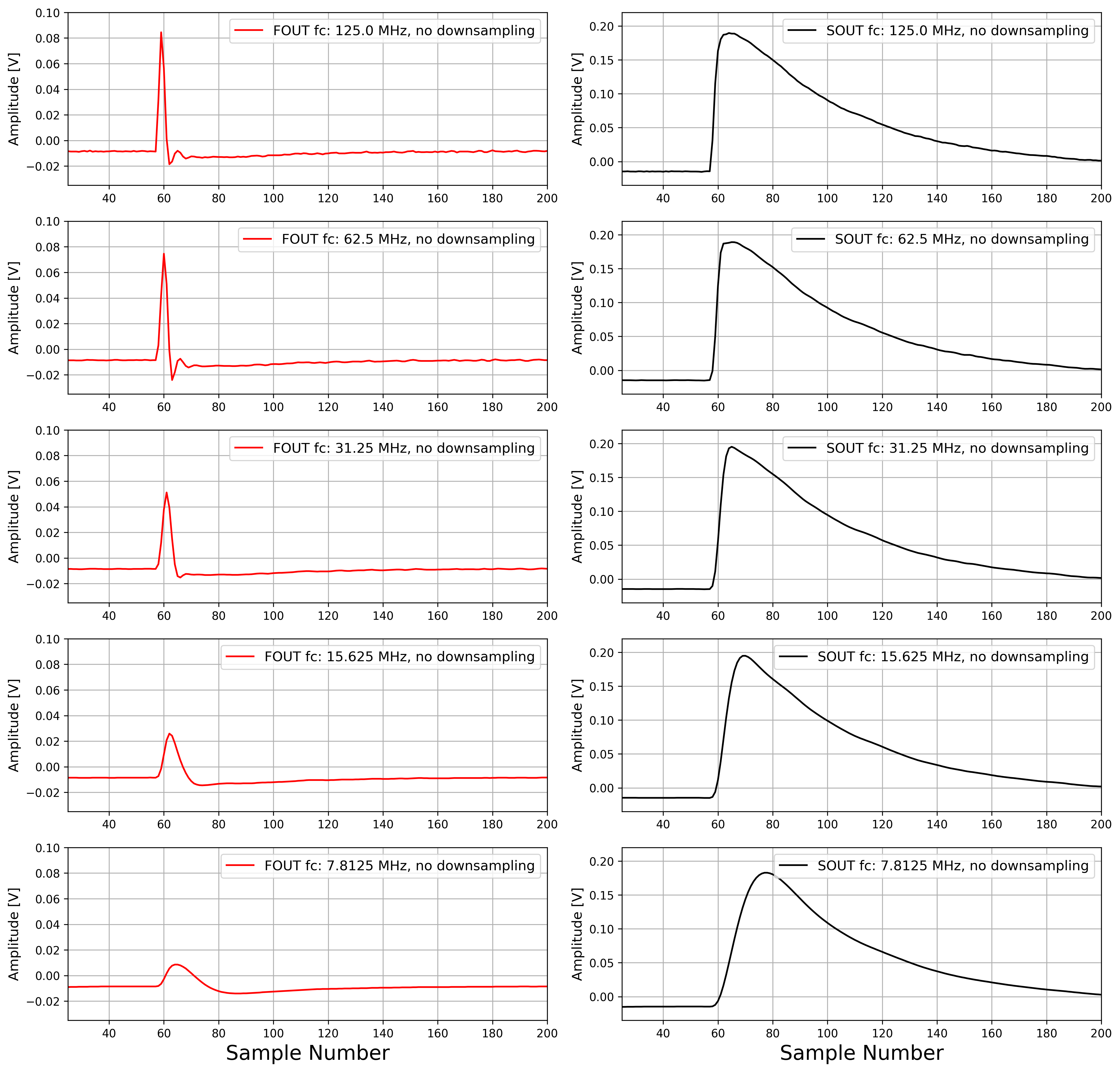}
    \caption{Example FOUT and SOUT waveforms re-filtered using a second order Butterworth filter implemented with the Python \emph{scipy} \cite{SciPy} package at various corner frequencies and 14-bit resolution. Downsampling of the waveforms is not performed for this figure. This is used in part to emulate the analog integration option for calculating input features.}
    \label{fig:nw_variousfrac_noDS}
\end{figure}

\begin{figure}[tb]
    \centering
    \includegraphics[width=0.8\textwidth]{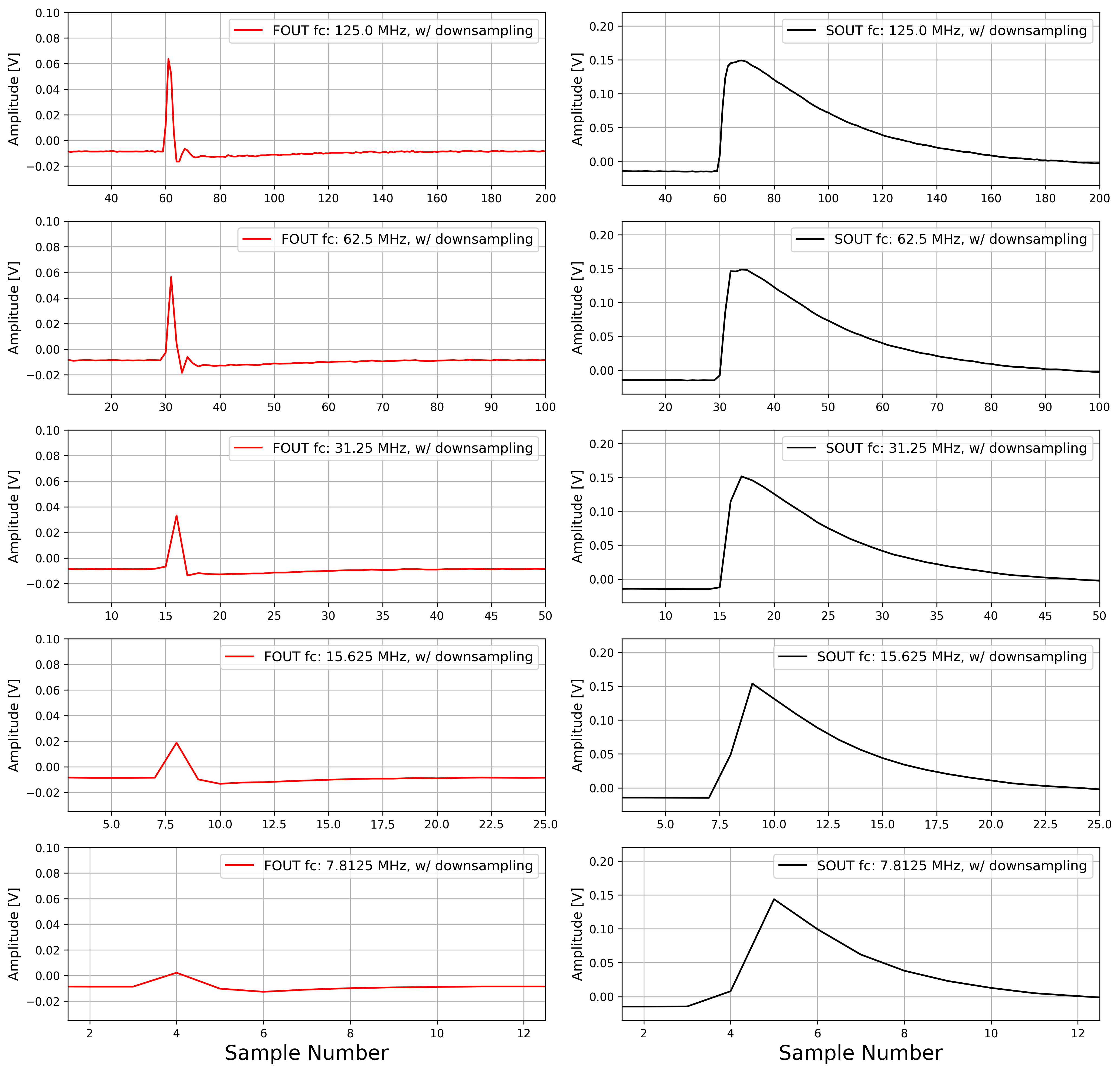}
    \caption{Example FOUT and SOUT waveforms re-filtered using a second order Butterworth filter implemented with the Python \emph{scipy} \cite{SciPy} package at various corner frequencies and 14-bit resolution. Downsampling of the waveforms is performed for this figure to equal the Nyquist rate equivalent for the software filter's corner frequency.}
    \label{fig:nw_variousfrac}
\end{figure}

In Section \ref{sec:hls4ml}, a distinction was made on the degeneracy in physically implementing the extraction of input features. The two options discussed include using dedicated analog circuits and digitizing the final values or using a waveform digitizer and calculating input features digitally in a dedicated digital signal processing block. We can look at the performance of the standard charge comparison (PSD ratio) method under both scenarios. For both possibilities, \SI{128}{\nano\second} was used for the partial integration window and \SI{1024}{\nano\second} was used for the total integration window. The actual number of samples corresponding to these time windows varies depending on downsampling for the second option.

For the analog circuit solution, the signal should be considered analog (having "infinite" resolution) until the final value is digitized. To emulate this signal chain, we re-filter the acquired waveforms from the dataset but do not downsample nor re-digitize them. This is reflected in Figure \ref{fig:nw_variousfrac_noDS}. The effective bandwidth (filter corner frequency) of the re-filtered waveforms should be interpreted as an effective front-end bandwidth. The bit-resolution, in comparison, still maps to the ADC. After partial and total integration values are calculated for these waveforms, the values are re-digitzed to the specified bit-resolution. In re-digitizing the final integration values, the nearest power of 2 that encompasses the numerical range of values is used to define the full-scale range. This same full-scale range is kept regardless of final targeted bit-resolution. 

\begin{figure}[tb]
    \centering
    \includegraphics[width=1.0\textwidth]{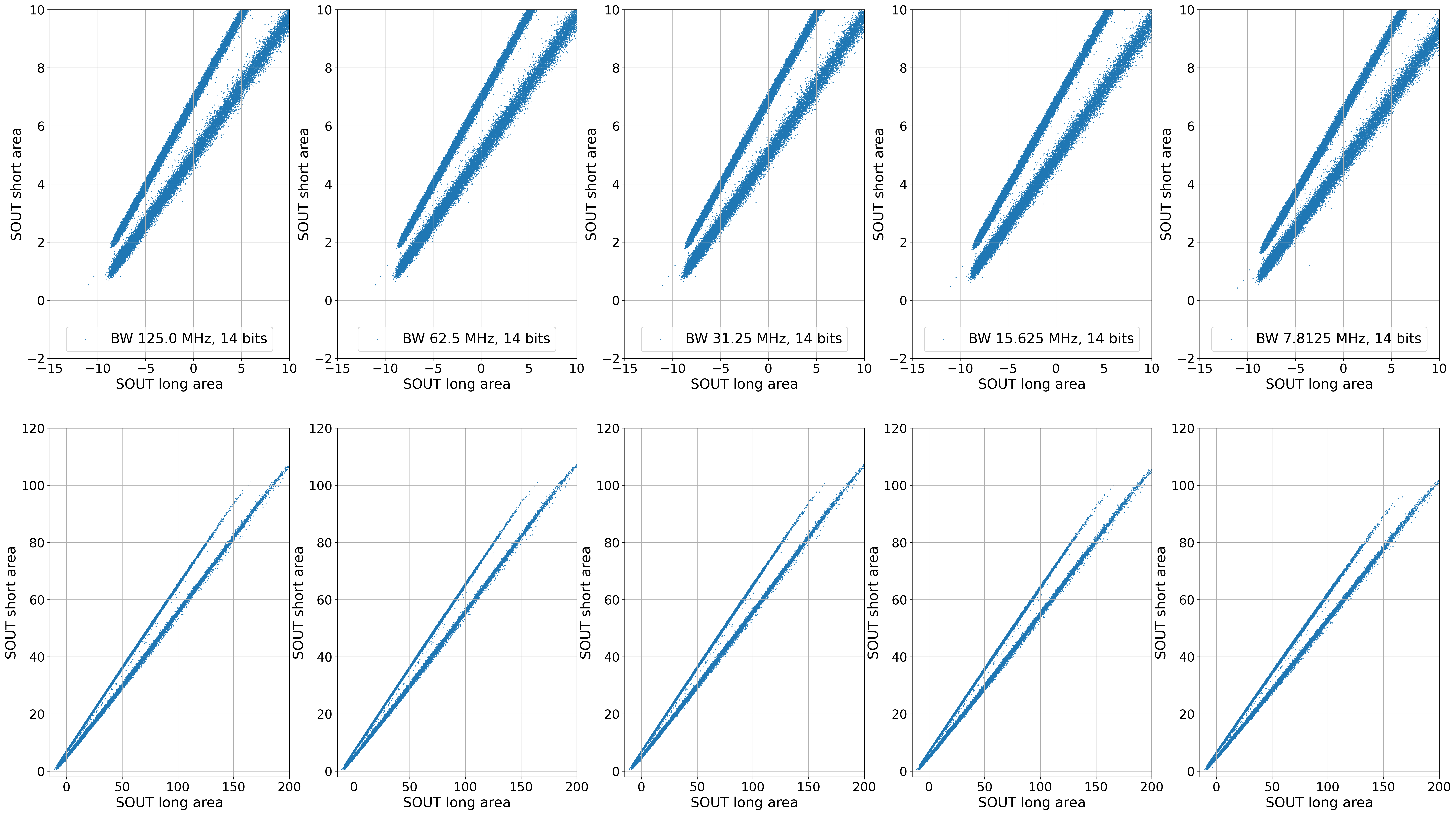}
    \caption{SOUT total vs. partial area is shown, both in units of \unit{\volt\us}, for various effective bandwidths based on re-filtering, but not downsampling waveforms for various filter corner frequencies. The figure explicitly shows the results for an ideal 14-bit resolution for the final calculated input feature values. Degradation of neutron/gamma band widths is not observed across all filter corners but band separation slightly decreases as the filter corner value decreases. The gamma band is above the neutron band.}
    \label{fig:area_plots_inputfeature_digitizer_best}
\end{figure}

For the waveform digitizer option, we performed re-digitization as well as downsampling to match the corner frequency of the software filter to emulate the expected signal chain. Examples for the FOUT and SOUT waveforms after this procedure are shown in Figure \ref{fig:nw_variousfrac}. Partial and total integration values are calculated on these resulting waveforms. Because we can define extra bits to store the final integration value, no re-digitization of the final integration values is needed for emulating this second option.

\begin{figure}[tb]
    \centering
    \includegraphics[width=1.0\textwidth]{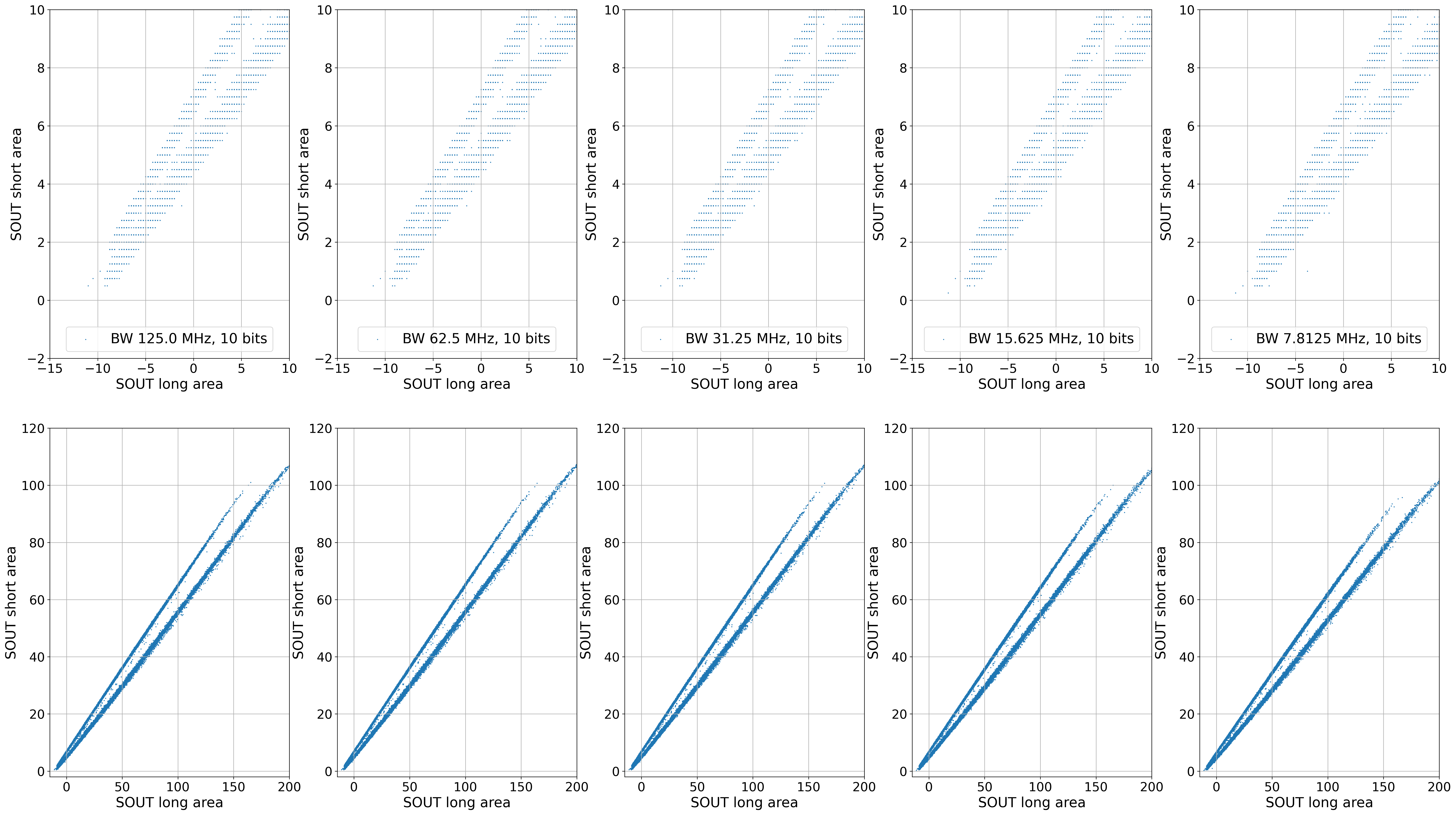}
    \caption{SOUT total vs. partial area is shown, both in units of \unit{\volt\us}, for various effective bandwidths based on re-filtering, but not downsampling waveforms for various filter corner frequencies. The figure explicitly shows the results for a non-ideal 10-bit resolution for the final calculated input feature values. Degradation of neutron/gamma band widths is observed across all filter corners, with band separation decreasing as filter corner value decreases. The gamma band is above the neutron band.}
    \label{fig:area_plots_inputfeature_digitizer_worse}
\end{figure}

\begin{figure}[tb]
    \centering
    \includegraphics[width=1.0\textwidth]{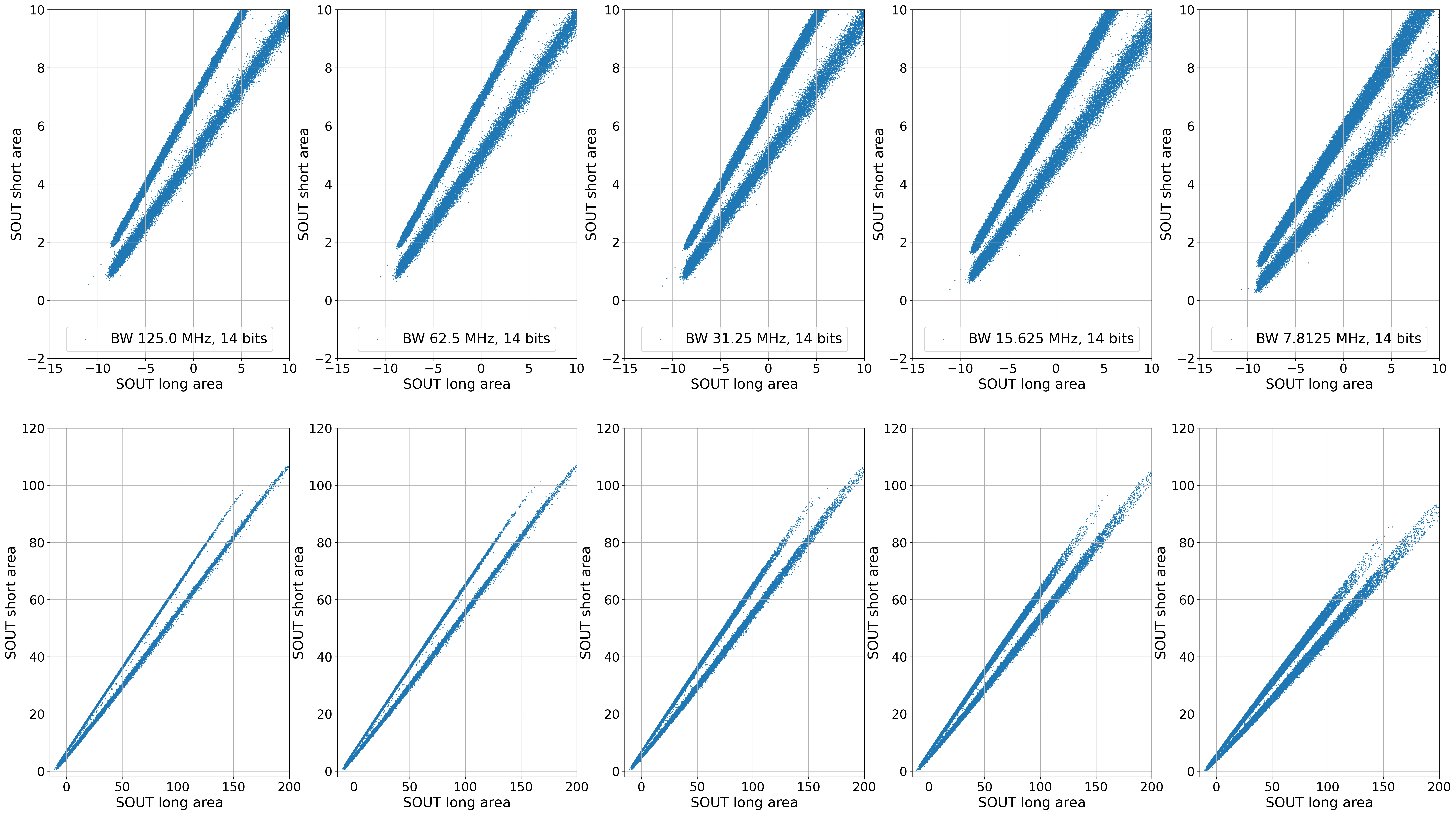}
    \caption{SOUT total vs. partial area is shown, both in units of \unit{\volt\us}, for various effective bandwidths based on re-filtering and downsampling waveforms for various corner frequencies and their corresponding Nyquist equivalent sampling rates. The figure explicitly shows the results for an ideal 14-bit resolution for the waveform samples. Degradation of neutron/gamma band separation and widths are observed at lower filter corner frequencies. However the effects are not as drastic compared to 8-bit waveform sample resolution in Figure \ref{fig:area_plots_waveform_digitizer_worse}. The gamma band is above the neutron band.}
    \label{fig:area_plots_waveform_digitizer_best}
\end{figure}

\begin{figure}[tb]
    \centering
    \includegraphics[width=1.0\textwidth]{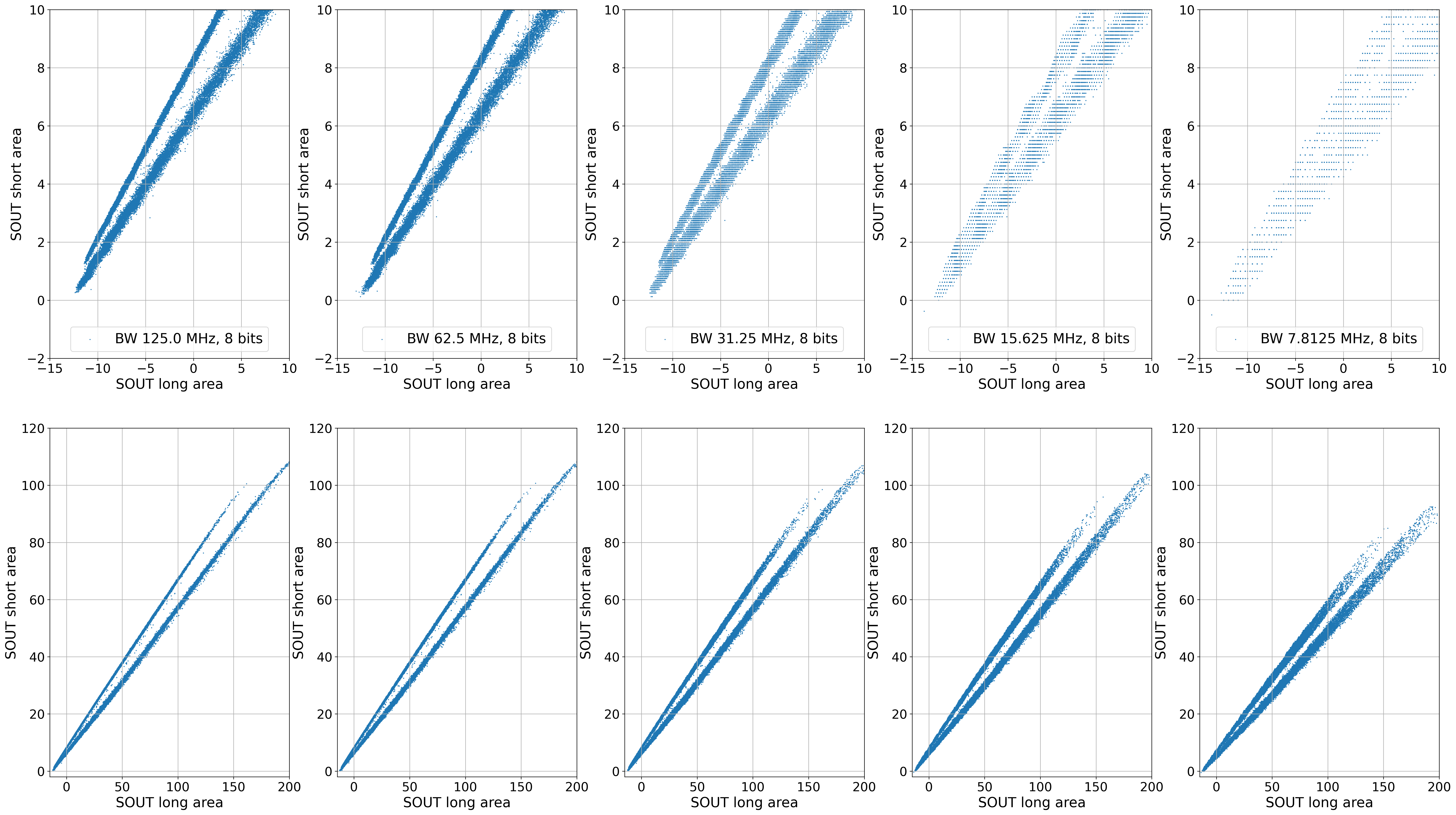}
    \caption{SOUT total vs. partial area is shown, both in units of \unit{\volt\us}, for various effective bandwidths based on re-filtering and downsampling waveforms for various corner frequencies and their corresponding Nyquist equivalent sampling rates. The figure explicitly shows the results for a non-ideal 8-bit waveform sample resolution. Significant degradation in neutron/gamma band widths are observed at lower filter corner frequencies with band separation degraded at all evaluated frequencies. The gamma band is above the neutron band.}
    \label{fig:area_plots_waveform_digitizer_worse}
\end{figure}

We can plot the calculated partial and total integration values for both possible solutions. Figures \ref{fig:area_plots_inputfeature_digitizer_best}, \ref{fig:area_plots_inputfeature_digitizer_worse} plot the calculated partial and total integration values assuming an analog integration and digitization of final values, with the former figure showing an ideal 14-bit resolution on the final values and the latter assuming 10-bit resolution. The bottom panel for both figures show a zoomed out view of all events from the dataset used for the study and the top panel shows a zoomed in view of the lowest energy events above the \SI{90}{\kilo\electronvolt}ee equivalent cut. At 10-bit resolution, a significant degradation of the band separation is observed at the lowest energies compared to 14-bits, where band separation of the original waveforms is maintained across different effective bandwidths. It should be noted that a slight degradation at the lowest considered bandwidth is visible. On the other hand, Figures \ref{fig:area_plots_waveform_digitizer_best}, \ref{fig:area_plots_waveform_digitizer_worse} plot the results for the second option. We see that even at 14-bit resolution, band separation and widths worsen with decreasing effective bandwidth. At 8-bit resolution, a significant degradation of possible band separation is observed even at the highest bandwidth considered. 

In general, it is clear that the standard charge comparison PSD metric is highly sensitive to the real-time implementation strategy and requires > 10-bit resolution and O(10) - O(100) MS/s for either possible implementation to maintain the ideal performance shown possible in \cite{Boxer_2023}. In that paper, it was demonstrated that gamma leakages of < $10^{-5}$ can be achieved while maintaining neutron efficiencies of 99.9\%. Maintaining that performance equates to the ability to maintain as close to the ideal band separation and widths at the lowest energy deposition events considered in this work.

As mentioned, for this ML model study, we are limited by dataset statistics to set gamma leakages at $10^{-3}$ but this should be treated as an upper bound as neutron efficiency can be maintained at lower leakage values. With a larger dataset, we can certainly set the performance metric at lower gamma leakage values. The results shown in Section \ref{sec:results} for both BDT and fcNN ML architectures should be considered from this point of view.

For the feasibility study, we only consider the waveform digitizer based approach to input feature calculation. For ML models with multiple input features, a waveform digitizer and calculation of values for each feature digitally is, in general, less complex than designing analog circuits for each feature and offers a more flexible path towards re-definition, if required. BDTs and fcNNs have different model hyperparameters and, thus, considerations for resource efficiency. As a result, we independently constructed evaluation pipelines for these two ML architectures to account for such differences. These are discussed in the proceeding subsections. Regardless of the architecture, the same four input features were calculated for each waveform, for each of the 35 sampling rate, bit-resolution pairs. The four input features are the SOUT total integration value (integration window of \SI{1024}{\nano\second}), SOUT partial integration value (integration window of \SI{128}{\nano\second}), FOUT peak amplitude, and SOUT peak amplitude. The exact integration windows used were selected to maintain compatibility with the various corner frequencies used for re-filtering of waveforms. The total integration window was changed from that used for labeling truth data in an attempt to decouple any potential biases in model training. Specifically, the partial integration window was selected to ensure that the window encompassed a sufficient area past the waveform peak amplitude to perform PSD even at the lowest filter corner value. The total integration window was set large enough to integrate on the full tail of the waveform. The input features were settled on after a dedicated importance study of various options. The ease for either analog or digital calculation of the features was kept in mind.

The final study parameter is the base unit used for mapping ADC bit resolution to fixed point format. While it seems that using \unit{\volt} as the base unit is the obvious choice to minimize input word length, the range of the input feature values and differences in ranges between the different input features plays a significant role in performance of fcNN models. For the BDT models, differences in model performance was also observed depending on choice of input feature value scale. For this reason, the choice was made to use \unit{\milli\volt} as the base unit for both ML model architectures despite the larger number of input bits required. For the area input features used, an area unit of \unit{\volt\micro\second} was assumed so that range of values matched to an order of magnitude with the amplitude features. A more thorough investigation studying the effect of input feature value scale and range can be undertaken but is outside the scope of this paper. In addition, to provide optimal performance capability for the fcNN models, the input features were normalized to mean 0 and set to a standard deviation of 1. The scaler values to accomplish this can be extracted from the trained fcNN model to keep the same normalization for new input feature values. This will need to be implemented as a pre-processing step for an actual real-time execution of fcNN ML model inferences. Despite this, we kept the input word lengths the same across ML algorithms so that resource usage estimates can more closely correspond to differences from the two ML architectures. 

%% file: Text/BDT.tex
For BDTs, the \emph{XGBoost} Python package provides a considerable amount of user optimization in finalizing a specific BDT model implementation. For this paper, we focused on the parameters that have the most effect on the final resource usage of the eFPGA and assumed fixed values for the other parameters. The main parameters that contribute to resource usage for the \emph{XGBoost} BDT model are max\_depth and num\_rounds. The scaling of FPGA resources with these two parameters has been studied and is defined in \cite{Summers_2020} and replicated here: 

\begin{equation} \label{eq:resource_BDT}
\mathrm{r} = k_{0} \cdot n_{b} + k_{1} \cdot n_{b} \cdot 2^{d}
\end{equation}

The terms $k_{0}$ and $k_{1}$ are unknown constants to be fitted for a given target, $n_{b}$ is the number of boosting rounds (num\_rounds) and $d$ is the max tree depth (max\_depth). Interested readers can refer to \cite{Summers_2020} for plots that visually show the scaling resulting from this equation. Although the \emph{conifer} implementation of the BDT allows selection of the quantization level for the input feature values, we assumed this was primarily determined by the ADC bit-resolution and appropriate scaling of the feature values as discussed. Similarly, the quantization level of the thresholds and the output raw scores were also selected respectively to account for the range of threshold values in the trained model consistent with the ranges of input feature values and to maintain overall performance of the model. Importantly, these parameters do not improve the performance of the BDT model past the baseline already set by the other parameters.  

In total, a four-dimensional parameter space to characterize performance on the neutron/gamma classification task remains: bit-resolution and sampling rate for the waveform digitizer (from which the input features are calculated) and the max depth and number of boosting rounds for the BDT model. Due to the exponential increase of resources with max depth, a fixed value of 3 was assigned after a preliminary look at model performance for max depth = 2 showed that significantly larger values for number of boosting rounds (> 200) are required to obtain similar performance to setting max depth = 3. In other words, the decrease in resource usage from using a max depth value of 2 is overwhelmed by the linear increase in resource usage from the number of boosting rounds value. As a result, we only focused on a final three-dimensional parameter space for this study.

%% file: Text/fcNN.tex
Much like \emph{XGBoost}, \emph{QKeras} for fcNNs has a large number of user configurable parameters. However, many more of the parameters have a non-trivial effect on the total resource usage of a trained model. In addition, unlike BDTs, there is no overall straightforward equation to describe the full parameter space's relationship to total resource usage. This is partly because the logic synthesis tool can optimize the exact physical implementation of the circuit. This can obfuscate resource scaling relationships due to degeneracies in implementing certain operations with DSP or LUT resources. Optimally, the decision is left primarily to the logic synthesis software unless specified as a directive/pragma by the user. We can obtain expected scaling of resources if iterating on a single parameter while keeping all others constant, as should reasonably be expected. However, to obtain a resource-efficient final model, a defined and ordered procedure for selecting values for the many contributing hyperparameters need to be defined. In total, there are nine considered hyperparameters that can affect the total resource usage: input feature bit resolution, output probability bit resolution, bit resolution for weights, biases (in theory, this can be specified per hidden layer), bit resolution of the activation function, total number of (and size of each) hidden layers, and target sparsity values for pruning aware training. A quantized tanh activation function is assumed for the hidden layer(s) and a sigmoid activation function for the output layer. The importance of quantization and the need to incorporate sparsity and pruning methods are described in \cite{coelho2020ultra}.

The order in which values are selected for these hyperparameters does matter. In general, the extent to which a model can be pruned/sparsified (i.e. the target sparsity value) without performance degradation partly depends on overall classification task complexity. Focusing on the neutron/gamma classification task at hand, it will depend strongly on the starting model size (number of hidden layers and size of each layer) as well as the input feature quantization level. So, we selected an ordered approach to defining values for the various hyperparameters.

An initial look was conducted by using the unmodified waveforms recorded with the test bed (14-bit and \SI{250}{\mega\sample\per\second}). The purpose was to narrow down the range of values for the overall hyperparameter space by finding lower bounds. From these initial results, a single hidden layer of size 8 was sufficient to ensure nearly unity neutron efficiency even with bit-resolution for the weights and biases set to 6 bits and for the activation function set to 10 bits. The number of epochs was set to 120 after the standard optimization against rate of loss function minimization. A target sparsity value of up to 50\% was found to still maintain neutron efficiency above 90\% in the initial look. These starting values for the hyperparameters provide anchor points when introducing input feature bit resolution and shaping. Instead of a large number of possible values for the nine parameters, we only accept a small upwards perturbation from these anchor values. For example, number of hidden layers was only allowed to vary between 1 and 2 and the size of the hidden layers could be either 8 or 16 for each hidden layer. Similarly, the number of bits for the weights and biases was allowed to be either 6 or 8, and for the activation function, either 10 or 12 if needed. Target sparsity values were allowed to vary between 10\% to 70\% and the number of epochs for training was fixed at 120. Unlike the number of boosting rounds parameter for BDTs, the number of epochs does not directly increase the model size.

%% file: Text/Results.tex
The results are considered independently for the two considered ML architectures.

%% file: Text/BDT_results.tex
\begin{figure}[tb]
    \centering
    \includegraphics[width=0.7\textwidth]{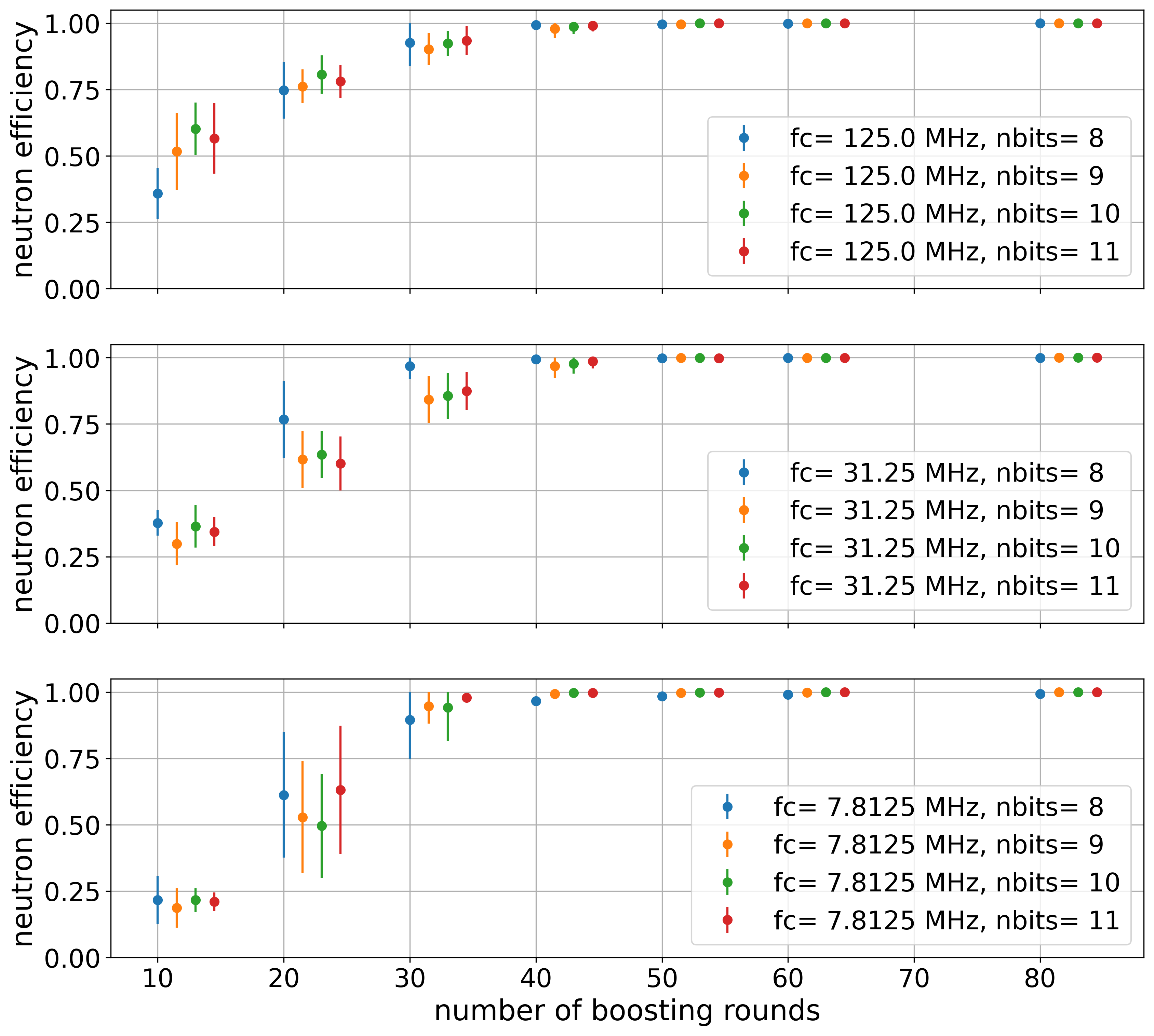}
    \caption{The mean neutron efficiency and $\pm1\sigma$ values are shown as a function of the $n_{b}$ value for bit-resolutions between 8 to 11 bits for BDT models. Each panel respectively shows the results for various evaluated filter corner frequencies. The filter corner frequency, also known as the \SI{3}{\decibel} point, represents the value at which the filter attenuates the input voltage signal by a factor of \(\frac{1}{\sqrt{2}}\). The efficiencies for nbits = 9 to 11 are plotted offset from the x value for nbits = 8, for visual clarity in the figure.}
    \label{fig:num_round_stat_var}
\end{figure}

We calculated the neutron efficiency value at a gamma leakage of 10$^{-3}$ for each of 35 pairs of bit-resolution and re-filtered corner frequency values (with downsampling) at several $n_{b}$ (number of boosting rounds) values. At all bit-resolutions and corner frequencies, it is possible to achieve > 95\% efficiency assuming an appropriate selection for $n_{b}$. Even with $n_{b}$ = 50, we demonstrate that an ADC with only 8-bit resolution and Nyquist sampling rates of approximately \SI{15}{\mega\sample\per\second} can still allow us to maintain excellent neutron efficiency at very low gamma leakage values in Figure \ref{fig:num_round_stat_var}. More generally, such a large parameter space for ADC specification where performance is maintained allows the final determination of design specs to be set by other system requirements for any specific target application.

One downside to selecting lower $n_{b}$ values, however, is that the range of statistical variations possible for the final BDT model start to become significant. This can be true despite the possibility to still achieve very good neutron efficiency values. 
A possible origin to these variations is that the information content in the input features dataset is more than can be taken advantage of when the $n_{b}$ value is too low. In this case, depending on initialization of thresholds during training, the model can achieve sub-optimal neutron efficiency values and, in general, display a high variance on final model performance.

To take this into account, a total of 20 trials were used as a compromise between required computation time and having sufficient sample size to calculate $\pm1\sigma$ neutron efficiency values for each $n_{b}$ value across filter corners. For each trial, we kept track of final neutron efficiency values achieved across a varying number of iterations of BDT model training and kept the model with the highest resulting value. The number of iterations varied between 200 for all $n_{b}$ values less than 50 to 100 at $n_{b}$ = 80. The results are shown in Figure \ref{fig:num_round_stat_var}. We observe significant variance on final neutron efficiency value despite this procedure. For example, at $n_{b}$ = 30, we see that, in general, the $\pm1\sigma$ neutron efficiency values are large compared to $n_{b}$ = 40. This trend gets progressively worse at even lower values. 

If we look individually at each bit-resolution performance at the various filter corner frequencies, we can observe some interesting differences across neutron efficiency results for $n_{b}$ = 10 to 40.
When we constrain the $n_{b}$ value to be low, the information content (discrimination power) of the input features dataset can also be insufficient. We see this manifest as lower resulting neutron efficiency values for the smaller filter corner frequencies. The standard procedure for software implementations is to simply increase the value used. For the case of targeting a heavily resource constrained hardware implementation, to do so is not as straightforward, since resource usage also increases.

Another important point is that neutron efficiency values will plateau after reaching some critical $n_{b}$ value and enter a diminishing returns regime. In this regime, for a hardware implementation, the resource requirements to implement the model increase but we only achieve marginally better performance. This behavior is easily observed in Figure \ref{fig:num_round_stat_var}. For this classification task, the plateau value is between 40 to 60, depending on filter corner frequency and bit resolution. In general, the value will also depend on both task complexity and the discrimination power of the selected/available input features. 

\begin{figure}[tb]
    \centering
    \includegraphics[width=0.8\textwidth]{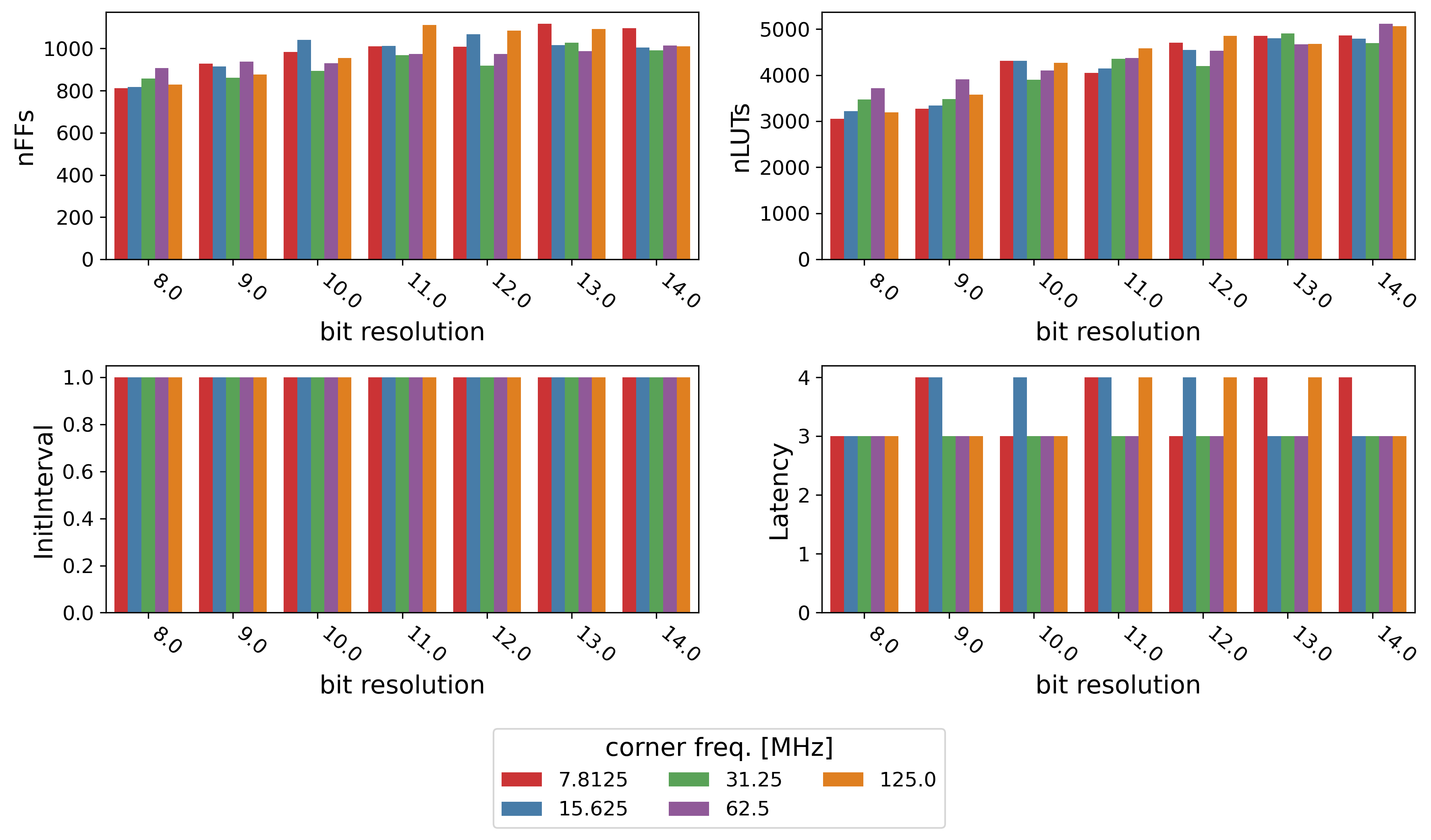}
    \caption{A detailed look at the nFFs and nLUTs required for a BDT model with $n_{b}$ = 50 across all evaluated bit resolution and filter corner frequency values. In addition, the returned latency and initiation interval values, in units of clock cycles, are given across all evaluated pairs.}
    \label{fig:resource_plots_BDT}
\end{figure}

Figure \ref{fig:resource_plots_BDT} histograms the number of flip-flops (nFFs) and number of look-up tables (nLUTs) for various bit resolution and corner frequency values as reported by Vivado post physical implementation. The selected $n_{b}$ value for the figure is 50 but the trends seen can be generalized. This value was selected to convey the required resources and resulting performance for a model with a neutron efficiency that has just plateaued. Models with a significantly lower fraction of required resources are possible if the neutron efficiency target is lowered. It is important to note that the large statistical variations of the neutron efficiency values do not cause similar effects on obtained resource usage, initiation interval and latency values. These are primarily set by the hyperparameters that define the BDT model. However, depending on the details of the final trained model, the synthesizer can optimize out small parts of the model which can manifest as small fluctuations of these values.

For testing purposes, a streamlined UART-based module was implemented as a wrapper to the model implementation so that input words can be written to the FPGA development board and results of an inference execution can be read back to verify performance on an actual Artix 7-series FPGA. The final nFF and nLUT values in the figure include the resources needed for implementing this module. A target clock frequency of \SI{100}{\mega\hertz} was used as a constraint. The required number of flip-flops is less than 3\% of the total available on the target device. The required number of LUTs across the parameter space is less than 25\% of the total available. 

While an initial increase in nFFs with bit resolution is observed, the numbers level off at higher bit resolutions. This is not the case for nLUTs. The required nLUTs increase with bit resolution and for each bit resolution, with filter corner frequency. No digital signal processor (DSP) nor block RAM (BRAM) resources are required for a BDT model implementation. The breakdown of the hardware implementation of BDT models and the expected mapping to resource usage is given in \cite{Summers_2020}. The minute deviations from the expected trends are most likely a result of optimizations made during logic synthesis and physical implementation by the Vivado synthesis tool. Across all bit resolution and sampling rate pairs, initiation intervals of 1 clock cycle were achieved. Similarly, latencies between 3 to 4 clock cycles of were achieved. The clock period for the study was fixed at \SI{10}{\nano\second} but in general can be lower, which would result in potentially lower absolute time values for initial intervals and latencies. 


%% file: Text/fcNN_results.tex
\begin{figure}[tb]
    \centering
    \includegraphics[width=0.7\textwidth]{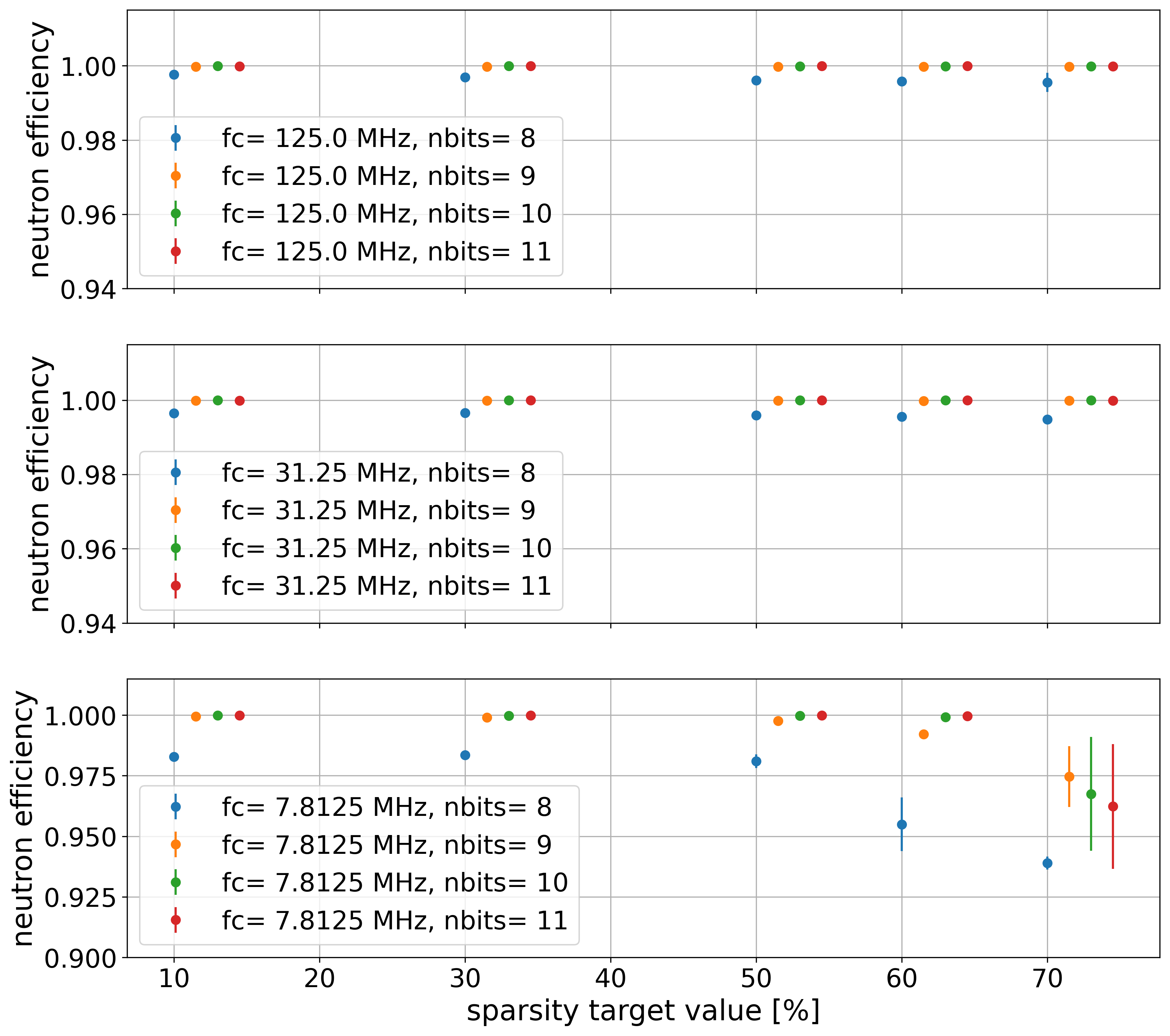}
    \caption{The mean neutron efficiency and $\pm1\sigma$ values are shown as a function of the target sparsity value for bit-resolutions between 8 to 11 bits for fcNN models. Each panel respectively shows the results for various evaluated filter corner frequencies. The filter corner frequency, also known as the \SI{3}{\decibel} point, represents the value at which the filter attenuates the input voltage signal by a factor of \(\frac{1}{\sqrt{2}}\). The efficiencies for nbits = 9 to 11 are plotted offset from the x value for nbits = 8, for visual clarity in the figure. The fcNN models all have one hidden layer of size 16 with the number of bits used for the weights and biases equal to 8 and equal to 12 for the activation function.}
    \label{fig:stat_var_sparsval_plot}
\end{figure}

We also calculated the neutron efficiency values at a gamma leakage of 10$^{-3}$ for each of the 35 pairs of bit-resolution and re-filtered corner frequency values (with downsampling) for trained fcNN models. Because of the larger parameter space, no exact comparable parameter to $n_{b}$ for fcNNs is possible. A single hidden layer of size 16, with the number of bits used for the weights and biases equal to 8 and equal to 12 for the activation function, is sufficient to maintain high neutron efficiency values across all 35 considered pairs. This is true even at large target sparsity values when performing pruning and quantization aware training. However, similar neutron efficiency values can also be obtained with a smaller hidden layer size and lower target sparsity values. We have used the larger size hidden layer model in Figure \ref{fig:stat_var_sparsval_plot}, which plots the observed statistical variance of fcNN models across a range of target sparsity values from 10\% to 70\%. As expected, with higher target sparsity values, the performance of the models at lower bit resolutions decreases, although overall neutron efficiency is still quite high. Significant statistical variance is not observed, unlike the BDT models. Some variance is seen at the largest target sparsity values considered. Unlike the BDT models, the fcNN models were not able to maintain the same level of neutron efficiency performance at 8 bits across all filter corner values even at a minimal target sparsity values of 10\%. Again, a total of 20 trials were used as a compromise between required computation time and having sufficient sample size to calculate $\pm1\sigma$ values for each target sparsity values. Unlike for the BDT models, only one training iteration was performed per trial because the number of epochs is set sufficiently high with no significant increase in resource usages.

\begin{figure}[tb]
    \centering
    \includegraphics[width=0.8\textwidth]{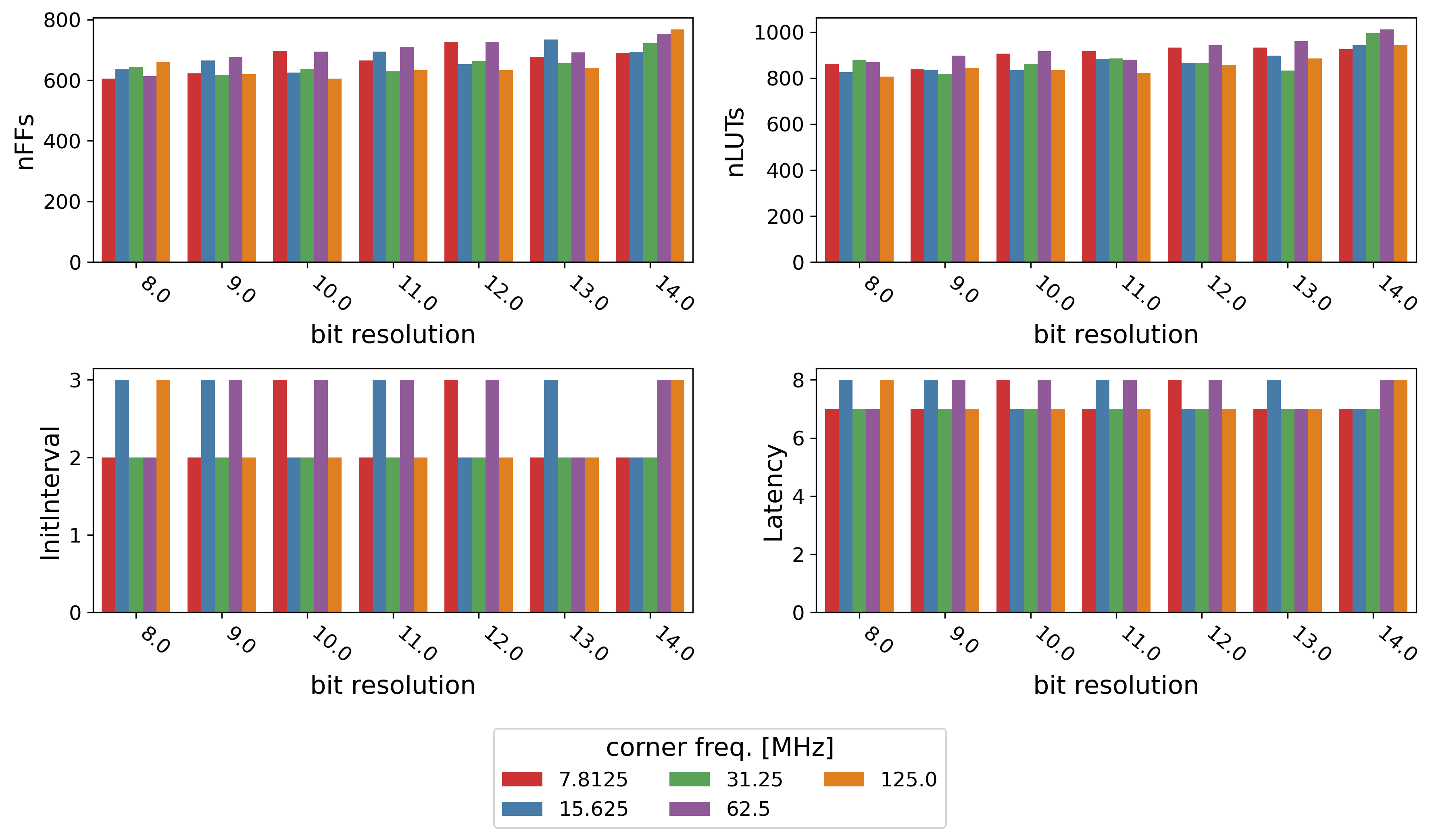}
    \caption{A detailed look at the nFFs and nLUTs required for fcNN models across all evaluated bit resolution and filter corner frequency values. The target sparsity values is set to 30\% and all fcNN models have one hidden layer of size 16. The number of bits used for the weights and biases is equal to 8 and equal to 12 for the activation function. In addition, the returned latency and initiation interval values are given, in units of clock cycles, across all evaluated pairs.}
    \label{fig:resource_plots_fcNN}
\end{figure}

Figure \ref{fig:resource_plots_fcNN} histograms the number of flip-flops (nFFs) and number of look-up tables (nLUTs) for various bit resolution and corner frequency values as reported by Vivado post physical implementation at a target sparsity values of 30\% for the above considered model. Like with the BDT model results, this value was chosen as representative of the edge of the plateau region. However, in this case, the decrease in neutron efficiency is minimal. The resources required to implement a fcNN model can be significantly reduced if the neutron efficiency target value is lowered, as is the case for the BDT models. We also see small fluctuations in the figure as a result of the synthesizer possibly optimizing out small parts of the model.

The required number of flip-flops is less than 3\% of the total available on the target device. The required number of LUTs across the parameter space is less than 10\% of the total available. Unlike BDT models, BRAM and DSP resources are required to store the activation function in look-up table format and to perform the required multiplications when executing an inference. The number of DSPs are not explicitly plotted but they vary between two to 11 across the hyperparameter values defined for the figure. This is at maximum, 12\% of the total available. Similarly, BRAM usage is reported as 0.5 of a single \SI{18}{\kilo\bit} block for all models. This is the smallest unit reported and the true usage of the full block is expected to be different across the evaluated parameter space. An identical clock frequency constraint of \SI{100}{\mega\hertz} is specified. In addition, the same UART-based wrapper is used to send input words to the FPGA development board and read back the executed inference result to verify performance. Thus, resource usage numbers reported include this module as well and should be an identical constant number.  

Unlike the BDT models, resource usage for nFFs and nLUTs are relatively constant across the hyperparameter space specified. The exact reason can be extrapolated to how a hardware implementation is approached for the two ML architectures. Regardless, resources will scale with target sparsity values much like they do with $n_{b}$ for BDTs. The exact nFFs required for fcNN models are higher than for BDTs and the opposite is true for nLUTs. However, this needs to be considered against the additional DSPs and memory resource tiles required to implement fcNN models. The breakdown of the hardware implementation of fcNN models and general trends on resource usage is given in \cite{duarte2018fast}. Across all bit resolution and sampling rate pairs, initiation intervals between 2 to 3 clock cycles were achieved. Similarly, latencies between 8 and 9 clock cycles were achieved. The BDT models clearly offer lower initiation intervals and latencies compared to fcNN models. This is expected from the constraints in gating the streaming of data between input to hidden and hidden to output layers for a given inference execution.

%% file: Text/Conclusion.tex
We have conducted a dedicated study to explore how to specify and train resource-efficient fcNN and BDT models for deployment on eFPGA fabrics using a stand-in commercial Artix 7-series FPGA target and a hardware-aware co-design methodology. We compared the study results to the golden standard of using the charge comparison method. The data set for the study was acquired with a fanout board where the choice of a \SI{50}{\ohm} resistor in series with the SiPM dominates the single photon response shape. In this regime, we were able to show that an ADC with only 8-bit resolution and Nyquist sampling rates of approximately \SI{15}{\mega\sample\per\second} can still allow us to maintain excellent neutron efficiency at very low gamma leakage values when using either multiple input feature ML architecture. For the standard charge comparison method, with the waveform digitizer option, a $\geq$ 10-bit, O(100) \unit{\mega\sample\per\second} ADC has to be used to maintain comparable performance. With the analog integration option, a similar resolution ADC is required and sampling rate will be set by expected event rate requirements since only the final integration values are digitized. 

We have calculated the resource types and quantities required for both ML architectures to implement non-trivial, high performance models to target the task of neutron/gamma classification. While the exact resource quantities and types differ, the general trade-off between BDT and fcNN models is that no DSP nor block RAM resources are required for BDT models but much larger numbers of LUTs are required compared to fcNN models. The translation to exact trade-offs in total required area will be process node and foundry dependent. However, having two fewer resource types to incorporate into the eFPGA fabric will simplify its design and implementation. The lower latencies and initiation intervals for BDTs provide an additional benefit if trying to maximize prediction throughput as is desired for the target applications for this case study. Thus, these two sets of benefits guide the decision to focus on BDT models and design the eFPGA fabric accordingly for a future first test chip. This first test chip will need to be larger than the one described in \cite{gonski2024embedded} if $10^{-3}$ gamma leakage is required to be maintained. However, for an initial demonstration, relaxing this value to $10^{-2}$ and optimizing the selection of input feature quantization can lower the amount of LUTs required to deploy high performance BDT models for neutron/gamma classification on eFPGA fabrics. 